\documentclass[lettersize,journal]{IEEEtran}
\usepackage{amsmath,amsfonts}
\usepackage{algorithmic}
\usepackage{algorithm}
\usepackage{array}
\usepackage[caption=false,font=normalsize,labelfont=sf,textfont=sf]{subfig}
\usepackage{textcomp}
\usepackage{stfloats}
\usepackage{url}
\usepackage{verbatim}
\usepackage{graphicx}
\usepackage{cite}
\usepackage{xcolor}
\usepackage{bm}
\usepackage{multirow}
\usepackage{bbding}
\usepackage{booktabs}
\usepackage{float}

\begin{document}

\title{ELBO-T2IAlign: A Generic ELBO-Based Method for Calibrating Pixel-level Text-Image Alignment in Diffusion Models}

\author{Qin Zhou, Zhiyang Zhang, Jinglong Wang, Xiaobin Li, Jing Zhang\textsuperscript{*},~\IEEEmembership{Member,~IEEE,}\\Qian Yu, Lu Sheng,~\IEEEmembership{Member,~IEEE,} and Dong Xu,~\IEEEmembership{Fellow,~IEEE}
\thanks{*Corresponding author}
\thanks{Qin Zhou, Zhiyang Zhang, Jinglong Wang, Jing Zhang, Qian Yu, Lu Sheng are with the School of Software, Beihang University, Beijing 100191,
China (e-mail: zhouqin2023@buaa.edu.cn, 20377279@buaa.edu.cn, wjlzy@buaa.edu.cn, zhang\_jing@buaa.edu.cn).}
\thanks{Xiaobin Li is with the College of Computer and Cyber Security, Hebei Normal University, Hebei 050024, China (e-mail: lixiaobin@buaa.edu.cn).}
\thanks{Dong Xu is with the Department of Computer Science, The University of
Hong Kong, Hong Kong (e-mail: dongxu@hku.hk).}
\thanks{Manuscript received April 19, 2021; revised August 16, 2021.}}

\markboth{Journal of \LaTeX\ Class Files,~Vol.~14, No.~8, August~2021}%
{Shell \MakeLowercase{\textit{et al.}}: A Sample Article Using IEEEtran.cls for IEEE Journals}

\IEEEpubid{0000--0000/00\$00.00~\copyright~2021 IEEE}

\maketitle

\begin{abstract}
Diffusion models excel at image generation. Recent studies have shown that these models not only generate high-quality images but also encode text-image alignment information through attention maps or loss functions. This information is valuable for various downstream tasks, including segmentation, text-guided image editing, and compositional image generation. However, current methods heavily rely on the assumption of perfect text-image alignment in diffusion models, which is not the case. In this paper, we propose using zero-shot referring image segmentation as a proxy task to evaluate the pixel-level image and class-level text alignment of popular diffusion models. We conduct an in-depth analysis of pixel-text misalignment in diffusion models from the perspective of training data bias. We find that misalignment occurs in images with small-sized, occluded, or rare object classes. Therefore, we propose ELBO-T2IAlign—a simple yet effective method to calibrate pixel-text alignment in diffusion models based on the evidence lower bound (ELBO) of likelihood. ELBO-T2IAlign is training-free and generic: it requires no additional annotations, model retraining, or architectural modifications, and it can be directly applied to different diffusion backbones. Extensive experiments on zero-shot referring image segmentation, text-guided image editing, and compositional image generation verify that the proposed calibration improves pixel-text alignment across complementary downstream tasks.
\end{abstract}


\section{Introduction}
\label{sec:intro}
\IEEEPARstart{T}{he} rapid advancement of text-to-image (T2I) diffusion models has enabled them to generate unprecedented results from given texts. Recent studies have shown that text-conditional diffusion models can not only generate high-quality images but also encode text-image alignment information~\cite{quangtruong2023dd,wu2023datasetdm,wu_diffumask_2023}. This information can be used for various downstream computer vision tasks, including image classification~\cite{li_your_2023,clark_text--image_2023}, object detection~\cite{chen2023diffusiondet}, image segmentation~\cite{noauthor_diffusion_2024,xu_open-vocabulary_2023}, and image editing~\cite{hertz_prompt--prompt_2022,cao_2023_masactrl,orgad2023editing}. For instance, some research reveals that attention maps in the denoising network encode object semantics~\cite{wang_diffusion_2024,tang_what_2023}. By effectively reusing these attention maps, diffusion models can be applied to semantic segmentation~\cite{wang_diffusion_2024} and panoptic narrative grounding~\cite{li2024dynamic}. Other studies analyze the diffusion model's loss function and utilize the loss value for image classification~\cite{li_your_2023,clark_text--image_2023}. Additionally, text-image alignment information can also be applied to text-guided image editing~\cite{hertz_prompt--prompt_2022,tumanyan_plug-and-play_2023} and compositional generation~\cite{chefer2023attend,feng2023training,rassin2023linguistic}. These downstream applications implicitly assume reliable text-image alignment in the frozen diffusion model, which often fails in practice. Calibrating such misalignment is therefore important for alignment-dependent tasks rather than only for image generation.

\begin{figure}[t]
\centering
\includegraphics[width=1\linewidth]{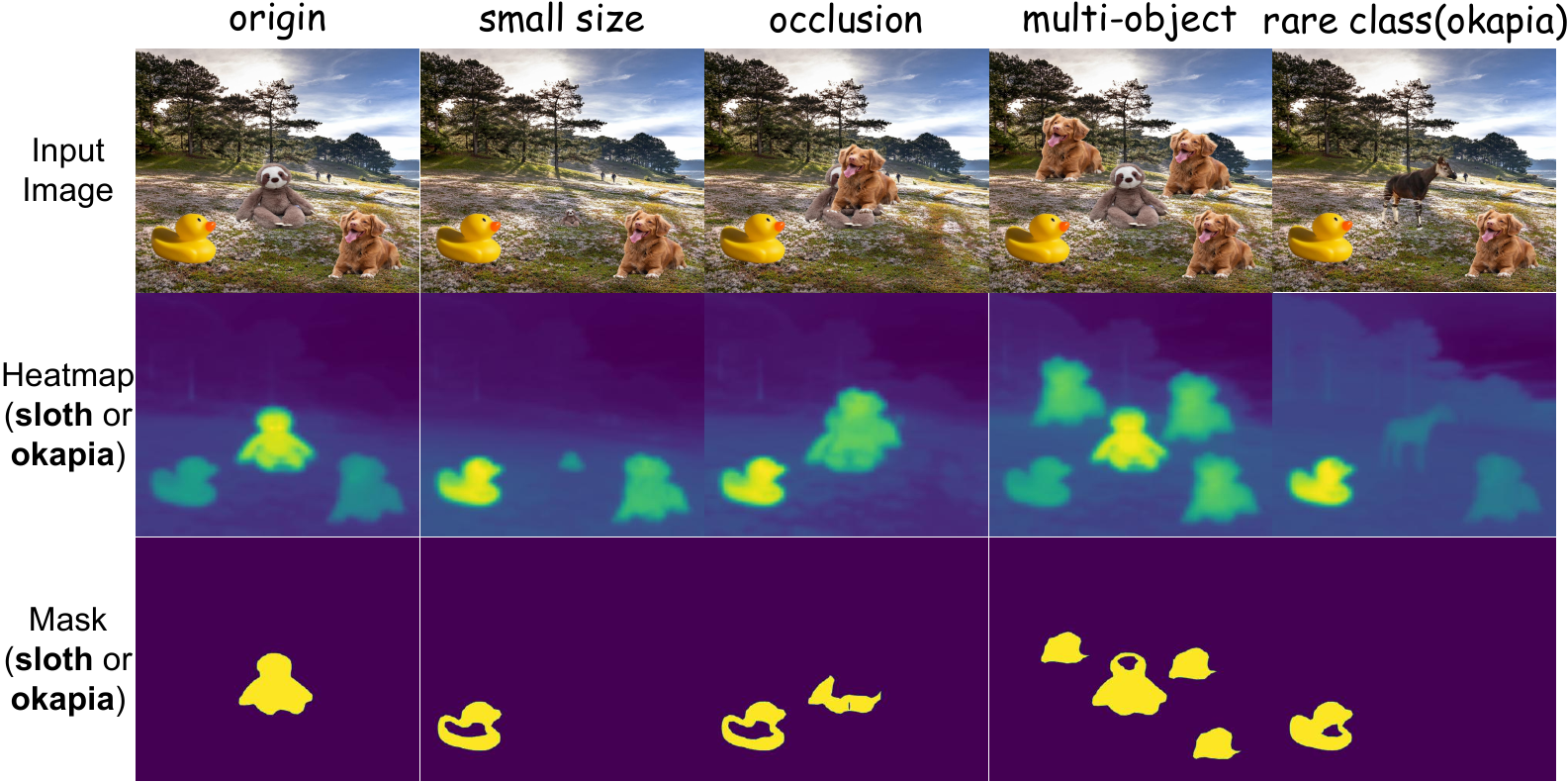}
\caption{
Illustration of how different data issues affect pixel-text alignment of diffusion models using toy data. The poor alignment will lead to inaccurate semantic heatmaps and masks.
}
\label{figure:toy_compare}
\end{figure}

In this paper, we examine the pixel-level image and class-level text alignment using zero-shot referring image segmentation as a proxy task, which can be effectively and quantitatively examined based on off-the-shelf semantic segmentation datasets. We find that one of the key reasons for the imperfect alignment between image pixels and textual classes stems from the biased dataset used for training the diffusion model. In the LAION dataset~\cite{schuhmann2021laion}, images are crawled from the internet. However, these internet images are naturally long-tailed due to Zipf's law of nature~\cite{reed2001pareto}, and they mostly contain common object classes rather than rare ones. The long-tailed issue in the LAION dataset has been validated by~\cite{parashar2024neglected,wen2024generalization}.\IEEEpubidadjcol Additionally, each image typically contains only a small number of classes. As a result, the text-image alignment is poorly established when multiple classes are present in a single image. We observe that when multiple classes appear in one image, the misalignment occurs with small-sized, occluded, or rare object classes. \figurename~\ref{figure:toy_compare} illustrates how different data issues affect pixel-text alignment in diffusion models using toy data (please see details in Section~\ref{sec:align_results}). The poor alignment will lead to inaccurate semantic heatmaps and masks.

To address these issues, we propose ELBO-T2IAlign, a simple yet effective training-free bias correction method to calibrate the pixel-text alignment. This generic approach works without needing to know the exact cause of misalignment, and works well across various diffusion model architectures. Specifically, we define an alignment score based on variational lower bound (ELBO) of the likelihood $p_{\theta}(x|c)$. Considering an image containing two object classes $c_1$ and $c_2$, we claim that $c_1$ has stronger semantics if $p_{\theta}(x|c_1)>p_{\theta}(x|c_2)$. We find that diffusion models always focus on classes that have stronger semantics, leading to higher activation values in cross-attention maps. In an image with multiple classes, most classes will have low activation values in cross-attention maps. To obtain ELBO of the log-likelihood, we simply use the diffusion loss functions, which are proved to be written as functions of ELBO~\cite{song_maximum_2021,kingma_understanding_2023}. Thus, we define the alignment score based on relative conditional diffusion loss values that encode information of ELBO of the log-likelihood. The alignment score corrects the cross-attention maps, ensuring that pixels of all object classes in an image have appropriate cross-attention activation values.

We conduct extensive experiments on three complementary tasks: zero-shot referring image segmentation for pixel-level localization, text-guided image editing for local controllability, and compositional generation for object-level text-image correspondence. The results demonstrate that our ELBO-T2IAlign approach consistently improves pixel-text alignment across different diffusion model architectures.

The contributions of this paper are three-fold:
\begin{itemize}
    \item We propose using zero-shot referring image segmentation as a proxy task to quantitatively evaluate the pixel-level image and class-level text alignment of popular diffusion models.
    \item We conduct an in-depth analysis of pixel-text misalignment in diffusion models from the perspective of training data bias.
    \item We propose ELBO-T2IAlign, a simple yet effective training-free generic method to correct pixel-text misalignment of diffusion models based on the ELBO of likelihood. Extensive experiments verify the effectiveness of our proposed correction approach on segmentation, image editing, and compositional generation.
\end{itemize}

\section{Related Work}
\label{sec:related_work}

\subsection{Diffusion Models}
Diffusion models have emerged as a new state-of-the-art (SOTA) in image and video generation, surpassing the Generative Adversarial Networks (GANs)~\cite{goodfellow2020generative}. These models generate samples by simulating a forward process that transforms data into noise and a reverse process that reconstructs data from noise. The foundational work, DDPM~\cite{ho_denoising_2020}, established the framework for diffusion models, while DDIM~\cite{song_denoising_2020} introduced a non-Markovian forward process that accelerates sampling. Latent Diffusion~\cite{rombach_high-resolution_2022} integrates diffusion models with latent variables, improving computational efficiency. Classifier-Free Diffusion Guidance~\cite{ho_classifier-free_2021} provides a conditional classifier that directs models to generate specific categories. ControlNet~\cite{zhang2023adding} introduces control signals for precise image editing. Flow Matching~\cite{papamakarios_normalizing_2021,lipman_flow_2022} uses optimal transport displacement interpolation to define the conditional probability paths, leading to better performance. Besides, research into diffusion models continues to explore new architectures, such as DiT~\cite{peebles_scalable_2023}, which uses Transformers instead of U-Net structures to enhance scalability and efficiency.

\subsection{Pixel-text Alignment Using Diffusion Models}

With the exploration and utilization of the attention mechanisms in diffusion models, recent research has focused on exploiting pixel-text alignment of the diffusion models through attention mechanism and applies the alignment nature to downstream tasks like semantic segmentation and image editing. For example, DiffSegmenter~\cite{wang_diffusion_2024} extracts cross-attention maps and self-attention maps, and combines them to achieve zero-shot semantic segmentation. OVAM~\cite{marcos-manchon_open-vocabulary_nodate} and DiffuMask~\cite{wu_diffumask_2023} utilize information from attentions to achieve pixel-level annotation of synthesized images. OVDiff~\cite{noauthor_diffusion_2024} leverages generated data to extract prototype features, thereby utilizing the perception capabilities of diffusion models. DiffSeg~\cite{tian_diffuse_2024} aggregates multi-layer self-attention maps to enable unsupervised segmentation with diffusion models. DiffPNG~\cite{yang_exploring_2024} uses the attention maps from diffusion as a basis, combined with the SAM~\cite{kirillov_segment_2023} model, to generate detailed masks. InvSeg~\cite{yang_exploring_2024} utilizes attention information to extract anchor points, optimizing the latent space and enhancing segmentation accuracy. Differently, PTP~\cite{hertz_prompt--prompt_2022}, Shape-Guided Diffusion~\cite{park2024shape}, CDS~\cite{nam_contrastive_2024}, and SAG~\cite{hong2023improving} utilize attention maps for controllable image editing. However, these methods overlook the training data bias inherent in diffusion models, resulting in suboptimal performance in pixel-image alignment.

\section{Preliminary}
\label{sec:Preliminary}

\subsection{Conditional Diffusion Models}
A conditional diffusion model is a conditional generative model $p_\theta(x|c)$ that approximates real conditional data distribution $q(x|c)$. In addition to observed data $x$, we have a series of latent variables $z_t$ for timesteps $t\in[0,1]$. Modern diffusion models process input in latent space. A condition encoder maps condition $c$ to text embeddings $\Gamma(c)\in\mathbb{R}^{L\times D_c}$, and an image encoder maps observed data $x$ to latent $x_0\in\mathbb{R}^{H_z\times W_z\times D_z}$. The model consists of a forward process and a reverse process. The forward process gradually adds noise to latent $x_0$, which is a Gaussian diffusion process with marginal distribution $q(z_t|x_0,c)$:
\begin{equation}
  z_t = \alpha_{t}x_0+\sigma_{t}\epsilon \ \ \text{where}\ \  t\in[0,1], \epsilon\sim\mathcal{N}(0,I),
  \label{eq:forward}
\end{equation}
where $\alpha_{t}$ and $\sigma_{t}$ are functions of $t$. The noise schedule (log signal-to-noise ratio) is given by $\lambda(t)=2\ln(\alpha_{t}/\sigma_{t})$. $\lambda(t)$ is strictly monotonically decreasing, thus gradually transforming the original $x_0$ into pure Gaussian noise $z_1$ in the forward process. The reverse process is given by inverting the forward process using a neural network $\epsilon_\theta(z_{t},t,c)$.

\subsection{ELBO Training Objective} The ELBO training objective is widely used in early diffusion models~\cite{ho_denoising_2020,song_denoising_2020}, which is equivalent to using KL divergence to directly compare the Gaussian distribution of the reverse process against forward process posteriors. The ELBO training objective of diffusion models is given by~\cite{kingma_variational_2021,song_maximum_2021}:
\begin{equation}
  ELB_\lambda(x,c)=\frac{1}{2}\mathbb{E}_{t,\epsilon}[-\frac{\mathrm{d}\lambda}{\mathrm{d}t}\left\|\epsilon_\theta(z_{t},t,c)-\epsilon\right\|_2^2],
\label{eq:elbo_formula}
\end{equation}
where $\lambda$ is the noise schedule given previously. Besides the ELBO objective, recent diffusion models also use different objectives or mixed objectives~\cite{songscore,lipman_flow_2022,ho_denoising_2020} instead of the exact ELBO\@. However, recent work~\cite{kingma_understanding_2023,esser_scaling_2024} showed that all diffusion training objectives can be formulated as weighted ELBO objectives, which establishes a bridge between recent diffusion losses and the earliest loss function of diffusion models.

\section{Method}
\label{sec:method}

\begin{figure*}[t]
    \centering
    \includegraphics[width=\textwidth]{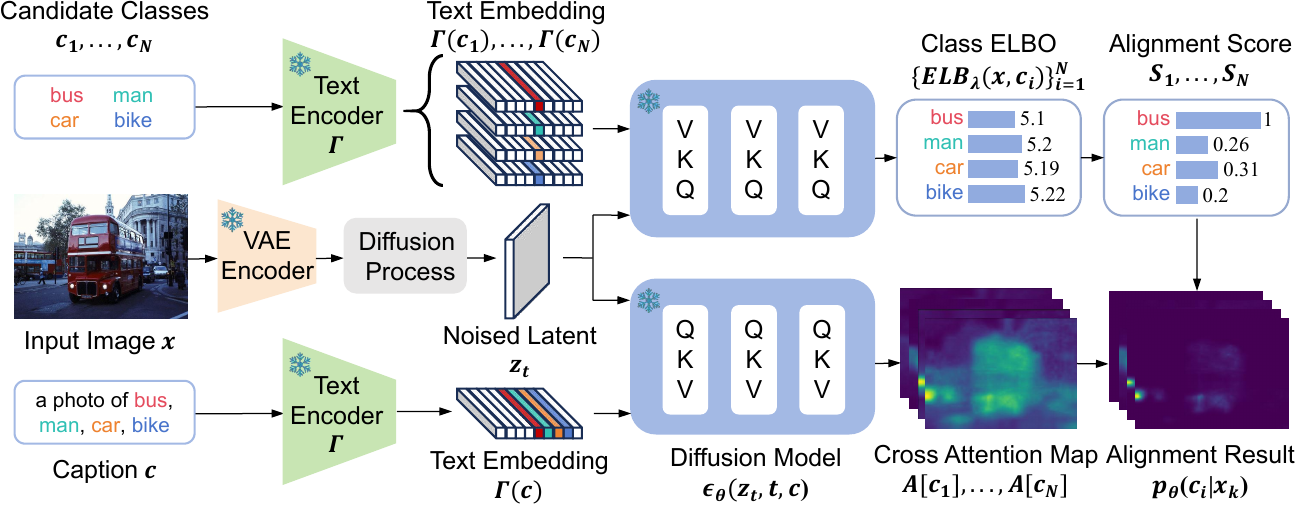}
    \caption{\textbf{Pipeline of ELBO-T2IAlign.}
    Given a pre-trained frozen diffusion model, we approximate rough pixel-text alignment using cross-attention maps. Then, we compute the ELBO of likelihood $p_\theta(x|c_i)$ for each class $c_i$. We define an alignment score based on ELBO, which is used to calibrate cross-attention maps. Segmentation masks are generated by thresholding $p_\theta(c_i|x_k)$.
    }
    \label{fig:pipeline}
\end{figure*}
In this section, we first define our task setting, analyze the misalignment issues, and outline our motivations. We then present our ELBO-T2IAlign method in detail.

\paragraph{Task definition} Given an image $x\in \mathbb{R}^{H\times W\times C}$ and candidate classes $\{c_1, \ldots, c_{N}\}$ in caption $c$, pixel-level image and class-level text alignment can be characterized by $p_{\theta}(c_i|x_k)$. Here, $c_i$ indicates the $i^{th}$ phrase that defines an object class, $x_k$ represents the $k^{th}$ pixel in image $x$, and $N$ is the number of classes. In an ideal alignment, if the class label of pixel $x_k$ is $c_i$, then $p_{\theta}(c_i|x_k)>p_{\theta}(c_{j\neq i}|x_k)$. We can generate segmentation masks directly using the approximated $p_{\theta}(c_i|x_k)$. To examine the pixel-text alignment, we employ zero-shot referring image segmentation as a proxy task. Zero-shot RIS aims to segment reference objects $c_i$ in image $x$ based on image caption $c$ without any training data, which can effectively reflect pixel-text alignment.

\paragraph{Analyses of misalignment}
The misalignment between image pixels and textual classes could be caused by several reasons. We take two classes as an example. According to likelihood and Bayes’ theorem, the posterior is,

\begin{equation}
    p_{\theta}(c_1|x_k)=\frac{p(c_1)p_{\theta}(x_k|c_1)}{p(c_1)p_{\theta}(x_k|c_1)+p(c_2)p_{\theta}(x_k|c_2)}.
\end{equation}
Suppose that the class label of $x_k$ is $c_1$, an ideal alignment means $p(c_1)p_{\theta}(x_k|c_1)>p(c_2)p_{\theta}(x_k|c_2)$. However, misalignment happens in at least three cases in practice:

\begin{enumerate}
    \item $p(c_1)\ll p(c_2)$, and $p_{\theta}(x_k|c_1)>p_{\theta}(x_k|c_2)$
    \item $p(c_1)\approx p(c_2)$, and $p_{\theta}(x_k|c_1)<p_{\theta}(x_k|c_2)$
    \item $p(c_1)<p(c_2)$, and $p_{\theta}(x_k|c_1)<p_{\theta}(x_k|c_2)$
\end{enumerate}
The first case stems from class imbalance with insufficient training data of $c_1$ (e.g., rare object classes or small objects present in an image). The second case indicates that even when different classes are approximately balanced, the wrong prediction could still happen due to class confusion between $c_1$ and $c_2$, which can occur due to similar visual appearances between $c_1$ and $c_2$, or when $x$ is an out-of-training-distribution sample for class $c_1$. The third case combines both class imbalance and confusion. In practice, the first case rarely occurs, since the model tends to assign higher likelihoods to classes that appear more frequently in the training data. Therefore, we mainly focus on the second and third cases.

\paragraph{Motivation}
\label{sec:motivation}

If the training set is balanced, the likelihoods of different classes $p_\theta(x|c_i)$ for a given image $x$ tend to be equal. Introducing two assumptions can ensure reasonably accurate segmentation results: (1) pixels are independent, implying that the image-level likelihood can be factorized into pixel-level likelihoods $p_\theta(x_k|c_i)$; and (2) the pixel-level likelihood within the target class region is larger than that in the background, enabling object localization. Leveraging the balanced data assumption, Bayes' rule implies $p_\theta(c_i|x_k) \propto p_\theta(x_k|c_i)$, which underpins the high-quality segmentation results of diffusion models.
However, real-world scenarios suffer from previously discussed issues: imbalanced training sets, out-of-training-distribution (OOD) problems, and class confusion. These issues cause the likelihood of specific classes to dominate, leading to abnormally large pixel-level likelihoods and subsequent missegmentation. Since diffusion models approximate pixel-level likelihoods via cross-attention maps and learn model likelihood via the ELBO objective, we propose leveraging ELBO as a measure of \textit{semantic strength}. This allows us to calibrate cross-attention maps by heuristically enhancing responses for classes with lower likelihoods, thereby effectively mitigating misalignment.

\paragraph{Overview}
An overview of our method is illustrated in \figurename~\ref{fig:pipeline}. Given a pre-trained frozen diffusion model, we encode the input image $x$ and caption $c$ using visual and text encoders, respectively. The visual features undergo noising through the diffusion process. Subsequently, we feed the textual features and noised visual features into the pre-trained denoising model. We approximate the rough pixel-text alignment using the cross-attention map. Then, we compute the ELBO of likelihood $p_\theta(x|c_i)$ for each class $c_i$ by passing individual classes as conditions. Based on the ELBO, we define an alignment score, which we use to calibrate the posterior approximated by cross-attention maps.

\subsection{Attention-Based Pixel-Text Alignment}
\label{sec:attention_based_align}

In diffusion models, obtaining the exact $p_{\theta}(c_i|x_k)$ is intractable. Previous methods approximate it either by the estimated likelihood~\cite{li_your_2023,clark_text--image_2023} or by using the cross-attention map~\cite{wang_diffusion_2024,tang_what_2023}.
In this paper, we approximate the rough pixel-text alignment $p_{\theta}(c_i|x_k)$ using the cross-attention map. Our method is applicable to all mainstream diffusion models since they all use attention mechanisms~\cite{NIPS2017_3f5ee243} to condition $x$ on $c$~\cite{hertz_prompt--prompt_2022,tumanyan_plug-and-play_2023,rassin2023linguistic}. For transformer-only diffusion models that rely on multi-modal self-attention (SD3, SD3.5, etc.), we split the whole attention map and categorize it into two groups: the text-image cross-attention map and the image-image self-attention map. We collect cross-attention map $A$ through one forward propagation of $\epsilon_\theta(z_{t},t,c)$:
\begin{equation}
    A=softmax(\frac{Q(z_t)K(c)^T}{\sqrt{d}})\in\mathbb{R}^{(H_z\times W_z) \times L},
    \label{eq:attention_map}
\end{equation}
where $d$ is a scaling factor, and $Q(z_t)$ and $K(c)$ are visual query and textual key matrices, respectively. Notice that (\ref{eq:attention_map}) only represents one cross-attention layer. In practice, since diffusion models may have multiple cross-attention layers with different resolutions, we resize all cross-attention maps to the same resolution and average them to produce the final $A$. Interaction between modalities primarily happens in cross-attention~\cite{rombach_high-resolution_2022,podell_sdxl_2023,peebles_scalable_2023,chen2023pixartalpha,esser_scaling_2024}, which computes the similarity between visual query and textual key matrices. Higher cross-attention scores yield larger activation values in $A$, signifying a stronger relationship between the image pixel and the text label. Therefore, we can approximate $p_\theta(c_i|x_k)$ by post-processing $A$. For each candidate class $c_i$, we first extract the corresponding columns in cross-attention map $A$ and average these columns to obtain $A[c_i]\in\mathbb{R}^{H_z\times W_z}$. Then, we apply min-max normalization and bilinear upsampling to $A[c_i]$, producing heatmap $H[c_i]\in\mathbb{R}^{H\times W}$. Next, we follow~\cite{wang_diffusion_2024,xu_mctformer_2023} to further enhance the heatmap $H[c_i]$ based on the self-attention map that encodes correlations between different pixels. Finally, we generate pixel-text alignment $p_\theta(c_i|x_k)$ using softmax. The algorithm is the same as Algorithm~\ref{alg:core} without Lines 3 and 7.

\begin{algorithm}[t]
    \caption{Pixel-Text Alignment Calibration with ELBO}
    \label{alg:core}
    \begin{algorithmic}[1]
        \REQUIRE Image $x$, caption $c$, candidate classes $\{c_1,\ldots,c_{N}\}$, timestep $t$ to collect attention maps
        \ENSURE Pixel-text alignment $p_{\theta}(c_i|x_k)$ that can directly be processed to segmentation masks
        \STATE Generate noised latent $z_t$ based on (\ref{eq:forward})
        \STATE Collect cross-attention map $A$ based on (\ref{eq:attention_map})
        \STATE Compute alignment score $\{S_i\}_{i=1}^{N}$ based on (\ref{eq:edp_core})
        \FOR{$i = 1$ to $N$}
            \STATE Extract corresponding columns of $c_i$ in $A$ and average them to obtain $A[c_i]\in\mathbb{R}^{H_z\times W_z}$
            \STATE Normalize: $A[c_i] \gets \text{min-max normalization}(A[c_i])$
            \STATE Pixel-text alignment calibration: $A[c_i] \gets A[c_i]^{\frac{1}{s_i}}$
            \STATE Class heatmap: $H[c_i] \gets \text{bilinear-upsample}(A[c_i])$
            \STATE Enhance heatmap $H[c_i]$ using self-attention map
        \ENDFOR
        \STATE Stack all heatmaps and apply softmax to generate pixel-text alignment $p_\theta(c_i|x_k)$ for $\{c_1,\ldots,c_N\}$
    \end{algorithmic}
\end{algorithm}

\subsection{Pixel-Text Alignment Calibration with ELBO}
\label{sec:elbo_enhance_method}
Based on previous discussions, we know pixel-text alignment $p_\theta(c_i|x_k)$ from cross-attention is inaccurate. One straightforward method to improve the result is explicitly modeling pixel-text alignment, but this approach involves redesigning training objective and training from scratch, which is costly. We propose to leverage ELBO of $p_\theta(x|c)$ from the pre-trained diffusion models to enhance $p_\theta(c_i|x_k)$ prediction. As mentioned in the motivation of our method, if an image contains two object classes, we claim that $c_1$ has stronger semantics if $p_{\theta}(x|c_1)>p_{\theta}(x|c_2)$. The relative magnitude of ELBO training objective reflects the semantic strengths, and can be used as a signal to instruct the calibration of the attention maps as well as the corresponding approximated $p_{\theta}(c_i|x_k)$.
We introduce an alignment score $S=\{S_i\}_{i=1}^{N}$ based on ELBO of $p_{\theta}(x|c_i)$ to identify the semantic strengths of each class in image $x$:
\begin{equation}
    S=\gamma^{norm(ELB_\lambda(x,c_1),\ldots,ELB_\lambda(x,c_{N}))}\in\mathbb{R}^{N},
\label{eq:edp_core}
\end{equation}
where $\gamma \in (0,1]$ is a hyper-parameter. We transform ELBO objective into alignment score $S_i \in [\gamma,1]$ by applying simple min-max normalization and point-wise exponent, while smaller $\gamma$ implies larger difference of alignment scores $\{S_i\}_{i=1}^{N}$. Intuitively, a higher alignment score $S_i$ indicates that $c_i$ has stronger semantic relevance, and also suggests that the model tends to misclassify pixels from other classes into this class. With the alignment score, we correct each $A[c_i]$ by applying an element-wise square root operation with a radical exponent $S_i$. Our calibration operation can be viewed as heuristically increasing the effective likelihood of classes with lower semantic strength, thus leading to a more reliable approximation of $p_\theta(c_i\mid x_k)$. The full algorithm is illustrated in Algorithm~\ref{alg:core}. Because all diffusion training objectives can be formulated as weighted ELBO, we can still estimate the ELBO of $p_\theta(x|c)$ by ``unweighting" different training objectives if we assume $\theta$ is well learned:
\begin{equation}
  ELB_{\lambda}(x,c)\approx\frac{1}{2}\mathbb{E}_{t,\epsilon}[\frac{1}{\omega(t)}\mathcal{L}_{\omega,\lambda}(x,t,c)],
  \label{eq:weight_ELBO}
\end{equation}
where $\mathcal{L}_{\omega,\lambda}(x,t,c)$ is the single-timestep training objective of diffusion models. We list $\omega(t)$ of popular training objectives in \tablename~\ref{tab:train_object}, and give the full derivation in the supplementary material. Furthermore, if we assume relative $\left\|\epsilon_\theta(z_{t},t,c_i)-\epsilon\right\|_2^2$ is stable among candidate classes across all timesteps, then the relative relationship of alignment scores will stay the same when using other objectives to replace $ELB_\lambda(x,c_i)$.

\begin{table}[t]
\caption{\textbf{$\omega(t)$ of popular diffusion training objectives.} Different objectives leverage neural network to estimate different functions, and use learned function to do sampling. See Section~\ref{sec:elbo_enhance_method} and supplemental material for more details and derivations.}
\label{tab:train_object}
  \centering
  \begin{tabular}{@{}lc@{}c@{}c}
    \toprule
    Objective & Estimated Function & $\omega(t)$ \\
    \midrule
    Score~\cite{songscore} & $\nabla_{z_t}\text{ln}q(z_{t}|x)$ & $-1/({\sigma_t^{2}}\cdot\lambda^{'}(t))$ \\
    Flow~\cite{lipman_flow_2022} & $u(z_{t}|\epsilon)$ & $-(\sigma_t^{2}\cdot\lambda^{'}(t))/4$ \\
    X~\cite{sohl-dickstein_deep_2015} & $x$ & $-1/(\lambda^{'}(t)\cdot e^{\lambda(t)})$\\
    Epsilon~\cite{ho_denoising_2020} & $\epsilon$ & $-1/\lambda^{'}(t)$\\
    Velocity~\cite{salimans_progressive_2021} & $\alpha_{t}\epsilon-\sigma_{t}x$ & $-(\alpha^2_{t}+\sigma^2_{t})^2/(\alpha^2_{t}\cdot\lambda^{'}(t))$\\
    \bottomrule
  \end{tabular}
\end{table}

\section{Experimental Results}
\label{sec:experiment}
In this section, we use zero-shot referring image segmentation as the primary quantitative proxy for pixel-level localization, and further examine text-guided image editing and compositional generation as complementary tests of local controllability and object-level text-image correspondence.

\subsection{Pixel-text Alignment Results}
\label{sec:align_results}
\paragraph{Datasets}
We evaluate our approach on the following five datasets.
PASCAL VOC 2012~\cite{everingham2010pascal} contains 20 classes with 1,449 validation images.
The PASCAL Context~\cite{mottaghi2014role} dataset includes 5,104 images for validation. We use the 59 most frequent classes, while the others are labeled as background.
COCO 2017~\cite{lin2014microsoft} contains 5,000 images for validation. We use the 80 object classes and set other classes to background. Ade20K~\cite{zhou2019semantic} contains 2,000 images for validation across 150 object and stuff classes. To further verify our method on \textbf{A}ttribute-\textbf{E}nriched \textbf{P}rompts, we synthesize a dataset named AEP\@. AEP contains 38 classes with attribute-enriched image captions (e.g. ``three baby rabbits and a pink modern wooden suitcase"). We collect 741 captions from DVMP~\cite{rassin2023linguistic}, generating and annotating images with FLUX~\cite{flux2024} and Grounded SAM~\cite{ren2024grounded}.

\paragraph{Implementation Details}
For datasets that do not contain image captions, we concatenate all candidate classes $\{c_i\}_{i=1}^N$ to produce caption $c$. For example, if an image contains cat and hot dog classes, the image caption will be ``a photo of cat, hot dog". Each candidate class is a noun phrase in the caption, and we generate pixel-text alignment $p_\theta(c_i|x_k)$ following Algorithm~\ref{alg:core}. To evaluate the result, we threshold $p_\theta(c_i|x_k)$, generate a background mask, and label other pixels with the highest probability class. Then, we compare the mask produced by our method with ground truth. We mainly use mean Intersection over Union (mIoU) as the evaluation metric. We use FP16 as default data type to reduce memory usage and accelerate inference. By default, we sample 20 timesteps to compute ELBO and set $\gamma=1/3$. As observed by DiffSegmenter~\cite{wang_diffusion_2024}, different cross-attention layers contain varying degrees of semantic information. Stronger semantic information is captured by lower-resolution cross-attention. We apply a weighted sum to all cross-attention layers, but assign different weights for different resolutions. We empirically use weights $15,10,1,1$ from the lowest to highest resolution. SDXL has 2 different resolutions, and SD3 has 1 resolution. We use weights $15,1$ for SDXL and simply average all layers for SD3. More details can be found in supplementary material.

\paragraph{Baselines} To verify the effectiveness of the pixel-text alignment enhancement by our method, we compare our method with the state-of-the-art diffusion-based segmentation methods, including DAAM~\cite{tang_what_2023}, OVAM~\cite{marcos-manchon_open-vocabulary_nodate}, DiffPNG~\cite{yang_exploring_2024}, Semantic DiffSeg~\cite{tian_diffuse_2024}, and DiffSegmenter~\cite{wang_diffusion_2024}. These methods can be easily adapted to our zero-shot referring image segmentation setting with few or no modifications. To ensure a fair comparison, the seed and text prompt are kept the same across different methods and other parameters are kept at their default values as provided. More details of baselines are in the supplemental material.

\paragraph{Quantitative Comparisons with Baseline Methods}
\tablename~\ref{tab:sota_result} presents a quantitative comparison of zero-shot referring image segmentation results between our method and other state-of-the-art diffusion-based methods. We evaluate mIoU on four benchmark datasets. Compared to baseline methods, our approach shows significant improvements across different datasets, which demonstrates its strong alignment capability. Our method only performs poorer than DiffSegmenter~\cite{wang_diffusion_2024} on the VOC dataset. We argue that since our method is designed to handle images that have multiple candidate classes, its effectiveness is less pronounced on the VOC dataset with fewer classes in an image.

\begin{table}[t]
\caption{Results of pixel-text alignment on four semantic segmentation benchmark datasets.}
  \label{tab:sota_result}
  \centering
   \resizebox{\linewidth}{!}{
  \begin{tabular}{c|cccc}
    \toprule
    \multirow{2}{*}{Method} & \multicolumn{4}{c}{\textbf{mIoU} $\uparrow$ }\\
     & VOC & Context & COCO & Ade20K \\
    \midrule
    DAAM~\cite{tang_what_2023} & 48.11 & 26.13 & 29.16 & 15.02 \\
    OVAM~\cite{marcos-manchon_open-vocabulary_nodate} & 47.45 & 25.17 & 25.72 & 11.68 \\
    DiffPNG~\cite{yang_exploring_2024} & 53.18 & 35.49 & 28.98 & 18.59\\
    DiffSegmenter~\cite{wang_diffusion_2024} & \textbf{68.68} & 39.47 & 38.88 & 22.88\\
    Semantic DiffSeg~\cite{tian_diffuse_2024} & 44.34 & 24.29 & 20.34 & 9.80\\
    ELBO-T2IAlign & \underline{68.15} & \textbf{43.61} & \textbf{42.15} & \textbf{26.17}\\
    \bottomrule
  \end{tabular}
  }
\end{table}

\begin{figure*}[t]
    \centering
    \includegraphics[width=\textwidth]{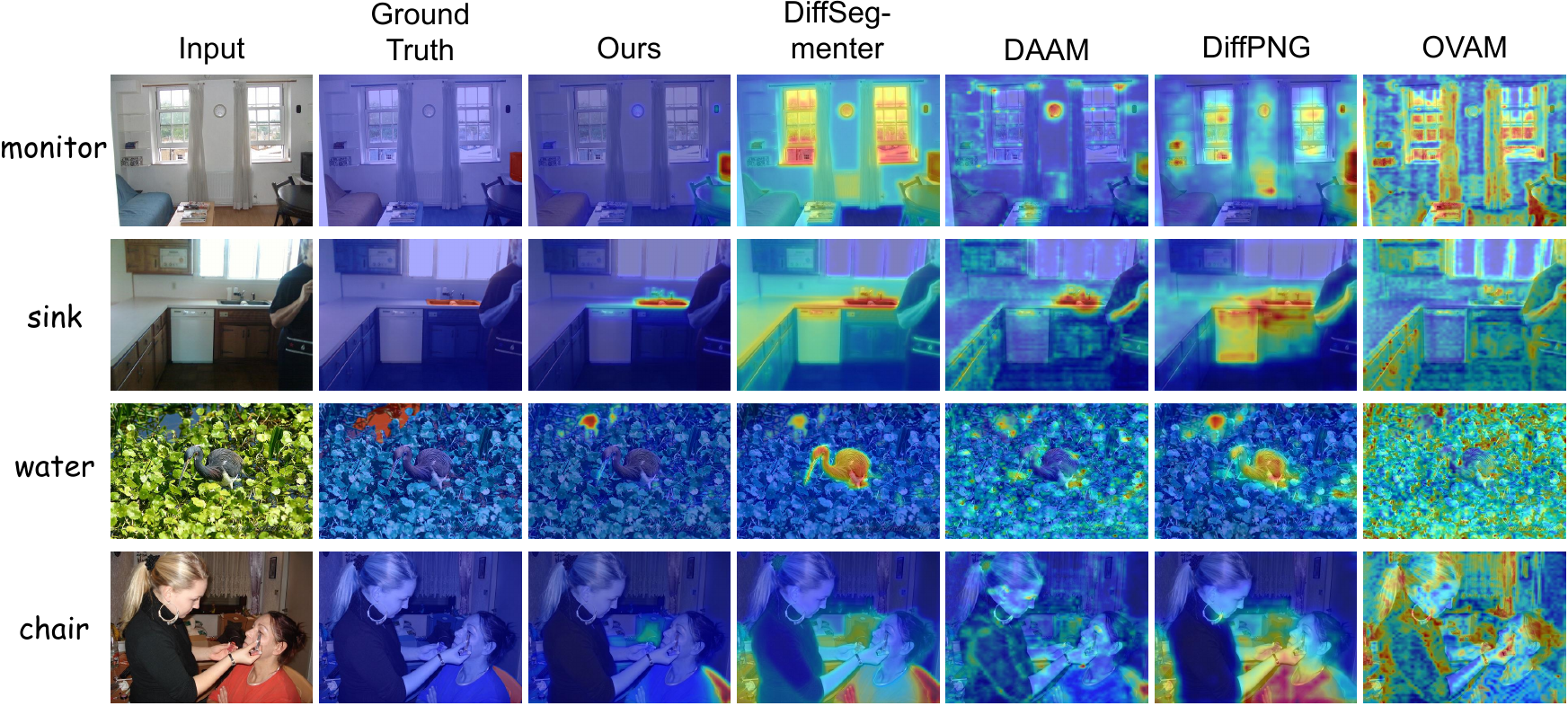}
    \caption{\textbf{Pixel-text alignment heatmaps} generated by our proposed ELBO-T2IAlign and the state-of-the-art diffusion-based segmentation methods.}
    \label{figure:main_compare}
\end{figure*}

\begin{figure*}[t]
    \centering
    \includegraphics[width=\textwidth]{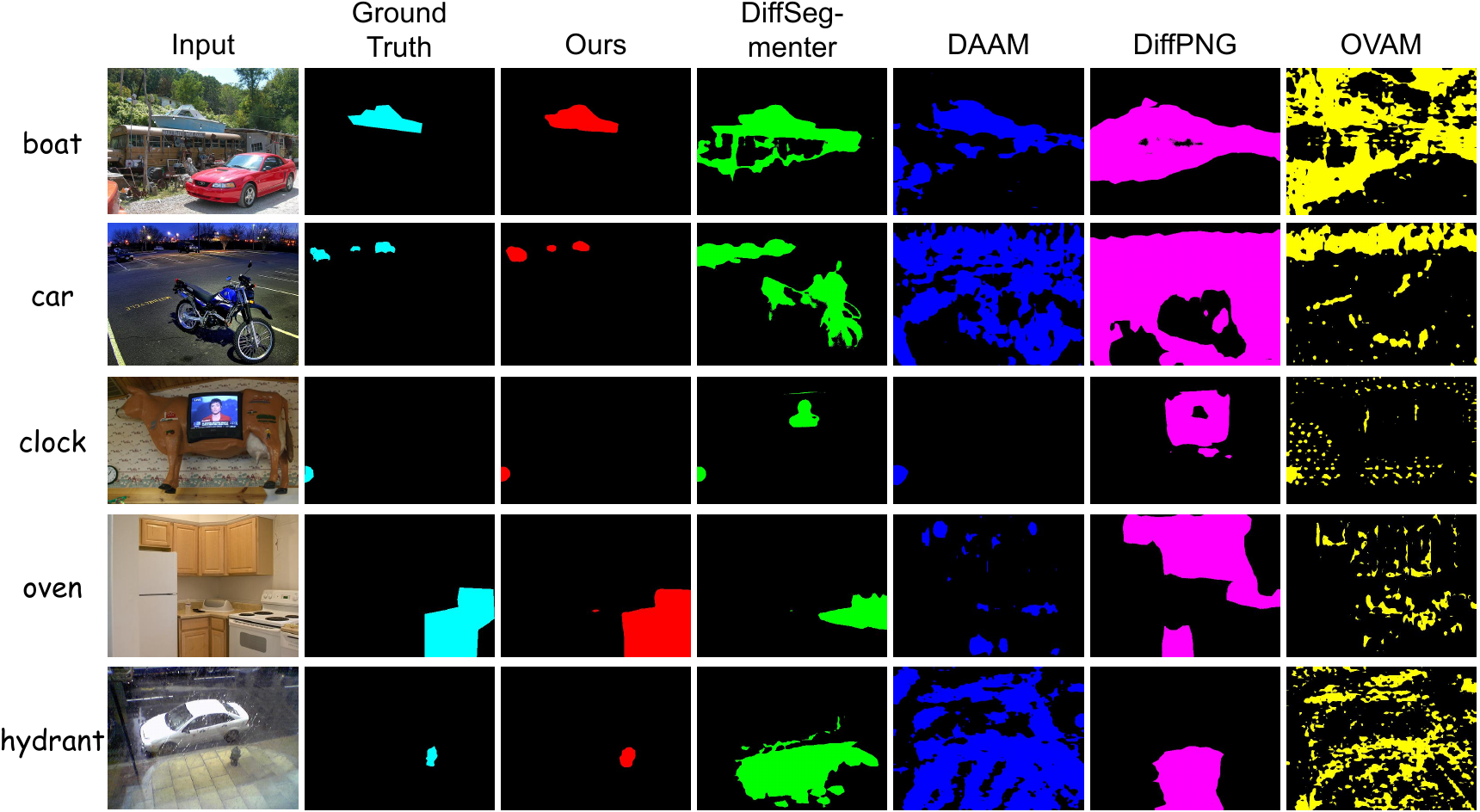}
    \caption{\textbf{Segmentation masks} generated by our proposed ELBO-T2IAlign and the state-of-the-art diffusion-based segmentation methods.}
    \label{figure:method_mask}
\end{figure*}

\paragraph{Qualitative Comparisons with Baseline Methods}
\figurename~\ref{figure:main_compare} illustrates the qualitative heatmap results comparing our method with other diffusion-based approaches. In the ``monitor" case, the monitor appears at the edge of the image and is only partially visible. In the ``sink" case, the sink occupies a very small portion of the image. In the ``water" case, water is a background element that is not easily recognizable compared to other objects, making intentional localization challenging. In the ``chair" case, the chair is obstructed by people and only the backrest is visible. The results indicate that our method achieves more precise localization and object contour capture for small objects like the sink, incomplete or obstructed objects like the monitor and chair, and less recognizable objects like water, compared to other methods. This demonstrates that our approach can better capture the semantic information of small or less prominent objects in multi-object scenes, reducing semantic confusion between different classes. We also present some segmentation masks generated by our method and other diffusion-based approaches in \figurename~\ref{figure:method_mask}.

\paragraph{Pixel-text Alignment Examination on Toy Data}

We conduct several toy experiments to examine how the semantic strength in an image affects the pixel-text alignment from pre-trained diffusion models. We observe at least four types of operations that reduce the semantic strength: small size, occlusion, multi-object, and rare class. Small size refers to proportionally scaling down an object, occlusion indicates occluding part of the object, multi-object refers to creating multiple copies of background objects, and rare class means replacing an object with a rarely seen object. We produce heatmaps and masks using Algorithm~\ref{alg:core} without Lines 3 and 7. As shown in \figurename~\ref{figure:toy_compare}, compared to the ``original" image, the pixel-text alignment of the targeted sloth plushie is weakened when its size is reduced, when occlusion occurs, or when multiple irrelevant objects are present. If we replace the target sloth plushie with a rare object (okapia), which has limited training data, the pixel-text alignment is also poor.

\paragraph{ELBO-T2IAlign Results of Various Diffusion Models}

\begin{table}[t]
\caption{Comparison results of \textbf{before and after calibration} using our ELBO-T2IAlign (ELB for short) across various diffusion models on four benchmark datasets.}
  \label{tab:various_model}
  \centering
   \resizebox{\linewidth}{!}{
  \begin{tabular}{cc|cccc}
    \toprule
    \multirow{2}{*}{Model} &\multirow{2}{*}{ELB}& \multicolumn{4}{c}{\textbf{mIoU} $\uparrow$ }\\
         &  & VOC & Context & COCO & Ade20K \\
    \midrule
    SD1.4 &  & 67.79 & 40.80 & 38.74 & 23.42\\
    SD1.4 & $\checkmark$ & \textbf{68.22} & \textbf{43.59} & \textbf{41.80} & \textbf{26.85}\\
    \hline
    SD1.5 & & 67.57 & 40.61 & 39.19 & 22.66 \\
    SD1.5 & $\checkmark$ & \textbf{68.15} & \textbf{43.61} & \textbf{42.15} & \textbf{26.17} \\
    \hline
    SD2 & & 63.71 & 37.39 & 39.60 & 22.65 \\
    SD2 & $\checkmark$ & \textbf{64.96} & \textbf{39.31} & \textbf{43.64} & \textbf{26.04}\\
    \hline
    SD2.1 & & 67.6 & 40.70 & 41.37 & 23.41 \\
    SD2.1 & $\checkmark$ & \textbf{69.04} & \textbf{42.41} & \textbf{44.53} & \textbf{26.32} \\
    \hline
    SDXL & & 57.72 & 33.28 & 34.89 & 17.87 \\
    SDXL & $\checkmark$ & \textbf{58.40} & \textbf{35.70} & \textbf{36.43} & \textbf{19.28} \\
    \hline
    SD3.5 & & 51.17 & 36.23 & 35.25 & 22.54 \\
    SD3.5 & $\checkmark$ & \textbf{51.74} & \textbf{37.27} & \textbf{37.31} & \textbf{25.01}\\
    \bottomrule
  \end{tabular}
  }
\end{table}

To demonstrate the generic nature of our method, we conduct experiments on diffusion models with different training objectives and architectures. The results are shown in \tablename~\ref{tab:various_model}. It is evident that across all tested models whose training objectives are different, ELBO-T2IAlign consistently improves performance. This consistent improvement suggests that our refined strategy contributes significantly to the model alignment capability, regardless of the specific training objective and architecture of the model.

\paragraph{ELBO-T2IAlign Results on Datasets with Attribute-enriched Prompts}

Since alignment score $S$ is the core of our method, we conduct experiments on generating $S$ with different strategies. Specifically, we are interested in whether using a whole noun phrase (e.g. ``a golden dog") is better than using a single noun (e.g. ``dog") in (\ref{eq:edp_core}). When computing alignment score $S$, does more accurate textual input lead to better results? We show the results on our AEP dataset in \tablename~\ref{tab:complex_prompt}. The results show that using the whole noun phrase as a candidate class is better than using a single noun. Using more accurate textual input is beneficial, which is consistent with recent methods that use a prompt enhancement module before generation.

\begin{table}[t]
\caption{Comparison results across various diffusion models on AEP dataset. We use different strategies to compute $S$ in (\ref{eq:edp_core}). \textbf{w/o ELB}: do not use ELBO for calibration, which implies $S_i$ is constant. \textbf{ELB}: use class name to compute ELBO for calibration. \textbf{ELB with AE}: use attribute-enriched noun phrase to compute ELBO for calibration.}
  \label{tab:complex_prompt}
  \centering
   \resizebox{\linewidth}{!}{
  \begin{tabular}{cc|c|c|c}
    \toprule
            Model& Setting & \textbf{mIoU} $\uparrow$ & \textbf{Prec} $\uparrow$ &\textbf{F1 Score} $\uparrow$ \\
    \midrule
    SD1.5 & w/o ELB & 58.70 & 68.77 &72.62\\
    SD1.5 & ELB  & 60.11 & 71.38 & 73.89\\
    SD1.5 & ELB with AE & \textbf{60.58} & \textbf{71.66}& \textbf{74.18}\\
    \hline
    SDXL & w/o ELB & 52.55 & 64.29 & 67.53  \\
    SDXL & ELB & 53.81 & 66.33 & \textbf{68.95}\\
    SDXL & ELB with AE & \textbf{53.82} & \textbf{66.96} & \underline{68.86}\\
    \hline
    SD3.5 & w/o ELB & 52.63 & 64.05 & 67.76  \\
    SD3.5 & ELB & 52.46 & \textbf{65.31} & 67.77\\
    SD3.5 & ELB with AE & \textbf{52.90} & \underline{65.13} & \textbf{68.16}\\
    \bottomrule
  \end{tabular}
  }
\end{table}

\subsection{Results on Image Generation and Editing}
\label{sec:additional_apps}

\paragraph{Compositional Image Generation}

\begin{table}[t]
\caption{{Quantitative compositional image generation results of our ELBO-T2IAlign and the baselines.}}
  \label{tab:elbo_generation_results}
  \centering
   \resizebox{\linewidth}{!}{
  \begin{tabular}{cc|ccc}
    \toprule
            \multirow{2}{*}{Method}& \multirow{2}{*}{Model} & \multicolumn{3}{c}{\textbf{CLIP score} $\uparrow$} \\
            & & A\&E & DVMP & ABC-6K  \\
    \midrule
    Baseline & SD1.5 & 31.45 & 32.33 & 32.73 \\
    A\&E~\cite{chefer2023attend} & SD1.5 & \textbf{33.09} & 32.24 & 32.48 \\
    SynGen~\cite{rassin2023linguistic} & SD1.5 & 31.83 & 32.18 & 32.45 \\
    ELBO-T2IAlign & SD1.5 & \underline{32.70} & \textbf{33.43} & \textbf{33.27} \\
    \hline
    Baseline & SD2.1 &  32.51 & 32.83 & 32.98  \\
    A\&E~\cite{chefer2023attend} & SD2.1 & \textbf{33.62} & 33.16 & 32.86 \\
    SynGen~\cite{rassin2023linguistic} & SD2.1 & 32.20 & 32.34 & 32.70 \\
    ELBO-T2IAlign & SD2.1 & \underline{33.24} & \textbf{33.49} & \textbf{33.30} \\
    \hline
    Baseline & SDXL &  33.61 & 34.27 & 33.22  \\
    ELBO-T2IAlign & SDXL & \textbf{33.72} & \textbf{34.45} & \textbf{33.37} \\
    \bottomrule
  \end{tabular}
  }
\end{table}

\begin{figure*}[t]
    \centering
    \includegraphics[width=\textwidth]{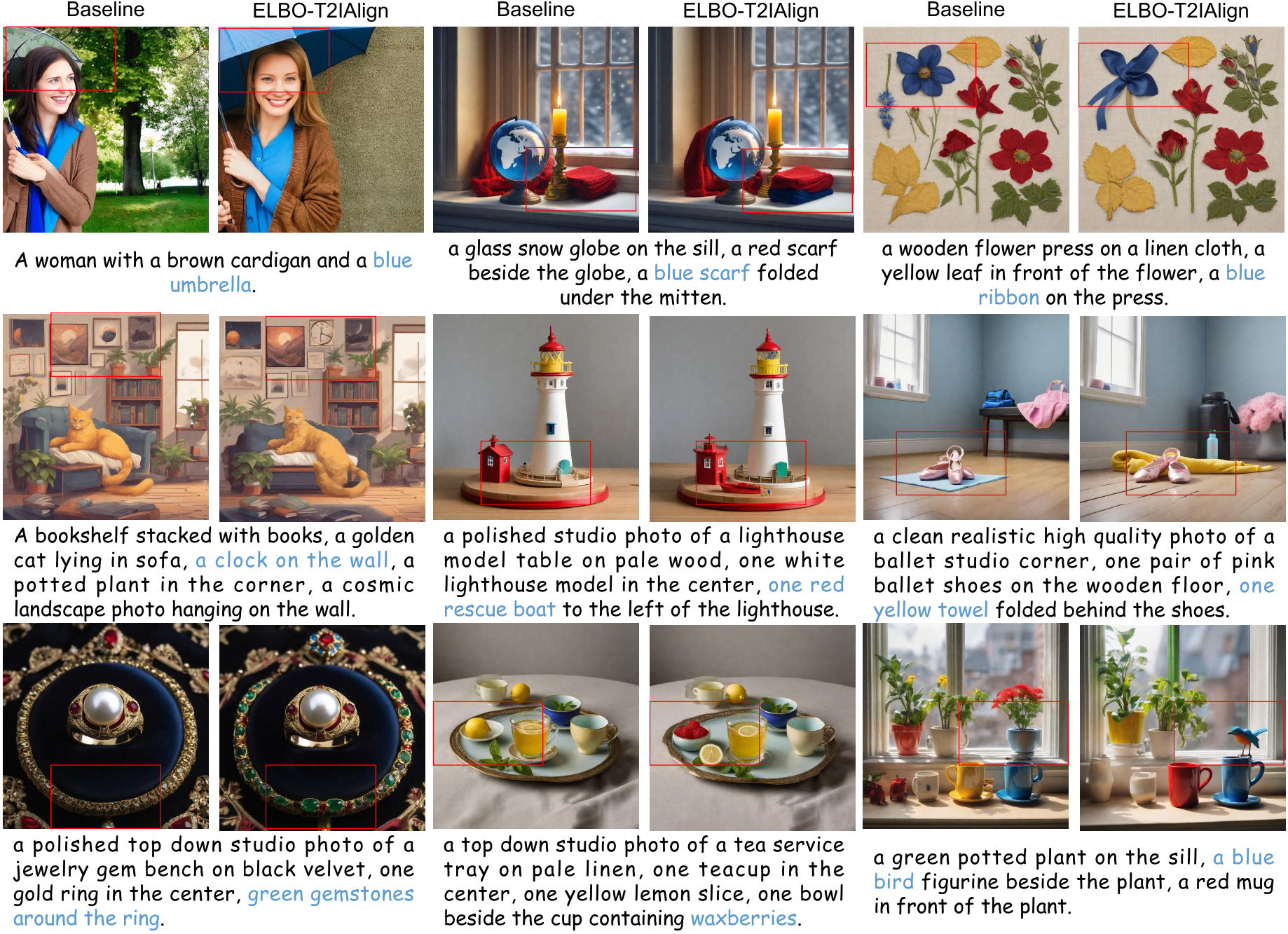}
    \caption{Qualitative compositional generation results of \textbf{before and after calibration} using our ELBO-T2IAlign.}
    \label{fig:gen_compare}
\end{figure*}

Our method can improve compositional generation using the proposed alignment score $S$. Given caption $c$, we extract entities and their visual attributes to obtain $\{c_i\}_{i=1}^N$. For each denoising step, we first estimate $\epsilon$ using $\epsilon_\theta(z_t,t,c)$. Then, we follow (\ref{eq:elbo_formula}) and (\ref{eq:edp_core}) to estimate $S$. Finally, we use $S$ for prompt reweighting~\cite{damian0815_compel,wang_diffusion_2024} to balance the semantics of $\{c_i\}_{i=1}^N$ and estimate $\epsilon$ with the reweighted prompt to denoise the latent. Specifically, when alignment score $S_{i}$ is low, we increase the weight of class $c_i$. Balanced semantics allow the generated image to better represent extracted entities, thus improving text-image alignment. Following a recent method~\cite{rassin2023linguistic}, we evaluate our method on the A\&E~\cite{chefer2023attend}, DVMP~\cite{rassin2023linguistic}, and ABC-6K~\cite{feng2023training} datasets using CLIP score. Compositional generation methods have two input settings: text-only input (A\&E~\cite{chefer2023attend}, SynGen~\cite{rassin2023linguistic}) and additional layout input (BeYourself~\cite{0Be}). Our method is designed for text-only input. We compare our method with text-only input methods in \tablename~\ref{tab:elbo_generation_results} and visualize some images in \figurename~\ref{fig:gen_compare}. A\&E and SynGen are coupled with the model architecture, so we do not implement them for the SDXL series or tune their hyper-parameters. The results show that our ELBO-T2IAlign method consistently improves text-image alignment over the baselines. We provide full details and more experimental results in the supplementary material.

\paragraph{Text Guided Image Editing}

\begin{figure}[t]
    \centering
    \includegraphics[width=1\linewidth]{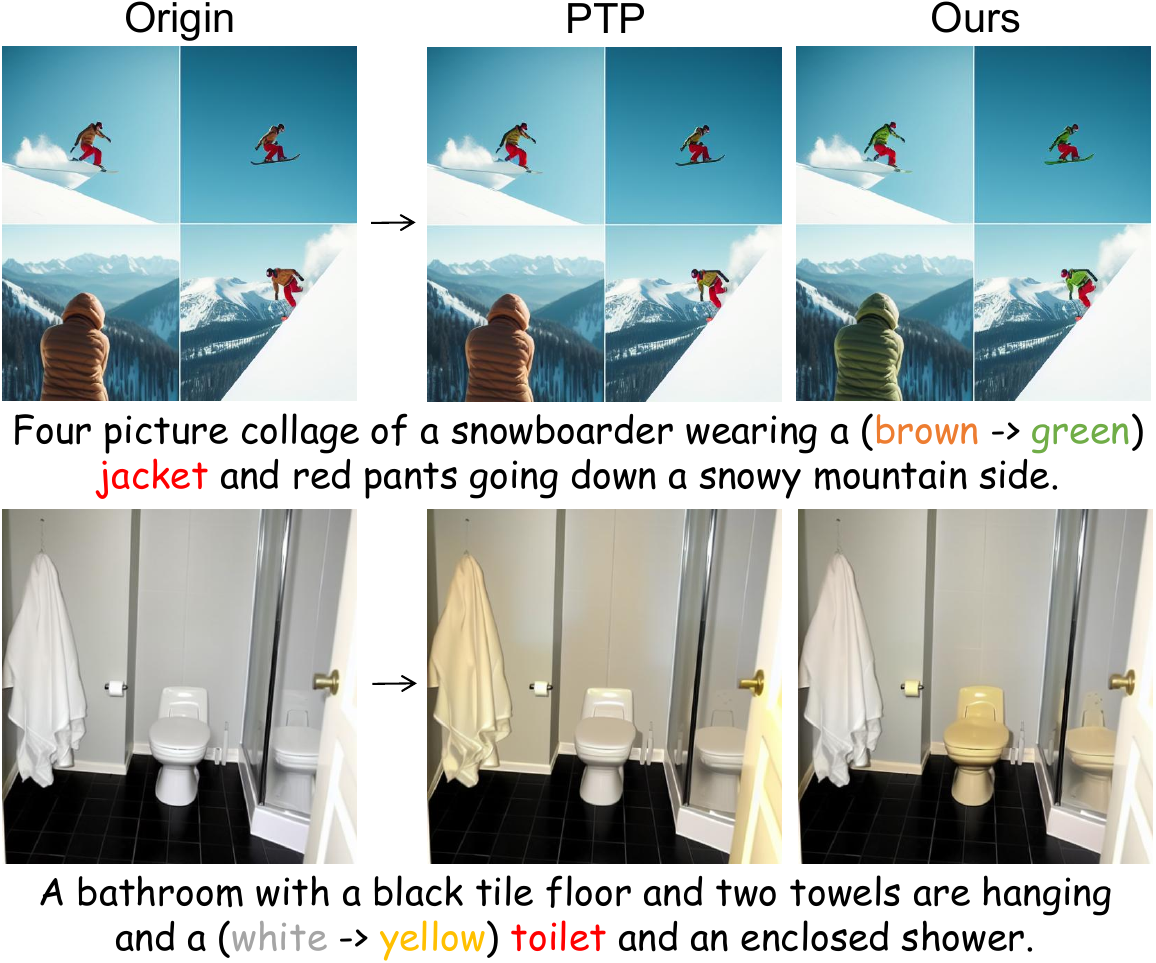}
    \caption{{Comparison results of image editing based on PTP~\cite{hertz_prompt--prompt_2022} \textbf{before and after calibration} using our ELBO-T2IAlign}.}
    \label{fig:edit_compare}
\end{figure}

Various diffusion-based image editing methods are based on cross-attention manipulation. Because ELBO-T2IAlign calibrates these maps, it can also improve local text-guided editing control. Specifically, we use PTP~\cite{hertz_prompt--prompt_2022} as baseline, which injects source text cross-attention to target text cross-attention to achieve editing. Given source text $c$, we extract entities and their visual attributes to obtain $\{c_i\}_{i=1}^N$. Then, we generate heatmaps of source text following Algorithm~\ref{alg:core}. Next, we scale these heatmaps based on source text cross-attention and use the scaled heatmaps to replace target text cross-attention. The visualization results are shown in \figurename~\ref{fig:edit_compare}. Our method can achieve editing more accurately with the scaled heatmaps. We give more results in the supplementary.

\subsection{Further Analyses}
\label{sec:ablation}

\paragraph{Analysis of Computational Resources Cost}

We primarily employ Python 3.11.9, PyTorch 2.5.1, transformers 4.51.3~\cite{wolf-etal-2020-transformers}, and diffusers 0.33.1~\cite{von-platen-etal-2022-diffusers} to implement our method on a single Ubuntu 20.04.3 LTS server with NVIDIA RTX 4090 graphics cards. Memory usage of our method is very close to model forward propagation, because we only perform forward propagation when computing ELBO and collecting attention maps. Our method uses approximately 7 GB of memory for Stable Diffusion v1/v2 and approximately 13 GB of memory for Stable Diffusion XL\@. For an image with 10 classes, our method takes around 4 seconds to generate all final masks on a single NVIDIA RTX 4090 graphics card with Stable Diffusion v1.5. The additional cost mainly comes from forward-only denoiser evaluations for estimating class-wise ELBO scores; no gradient computation, optimization, retraining, extra annotations, or architectural modifications are required. The timestep ablations show that the cost can be controlled by reducing the number of sampled timesteps when a faster setting is needed. If more GPU memory is available, our method can compute ELBO for all classes in a single batch to accelerate inference.

\paragraph{Effect of the Number of Timesteps to Compute ELBO}

To calculate ELBO, we sample timesteps evenly in the range $[0,1]$. In \tablename~\ref{tab:elbo_timestep}, we demonstrate the impact of sampled timesteps by varying $\text{ELBO Timesteps} \in \{1,5,20\}$ while keeping other hyperparameters unchanged. We find that as the number of sampling points increases, the mIoU on each dataset shows an upward trend. However, this increase is gradual, and inference time rises with more sampling points. Therefore, we balance these factors and choose $\text{ELBO Timesteps}=20$ as the default number of sampling steps for ELBO computation in our main experiment. We give more results in supplementary material.

\begin{table}[t]
\caption{ELBO-T2IAlign results on four benchmark datasets when using different numbers of timesteps to compute ELBO.}
  \label{tab:elbo_timestep}
  \centering
   \resizebox{\linewidth}{!}{
  \begin{tabular}{cc|cccc}
    \toprule
    \multirow{2}{*}{Method} &\multirow{2}{*}{Steps}& \multicolumn{4}{c}{\textbf{mIoU} $\uparrow$ }\\
         & & VOC & Context & COCO & Ade20K \\
    \midrule
    SD1.5 & 1 & 67.35 & 40.58 & 39.80 & 24.58 \\
    SD1.5 & 5 & \underline{68.15} & \underline{42.28} & \underline{41.36} & \underline{24.93} \\
    SD1.5 & 20 & \textbf{68.15} & \textbf{43.61} & \textbf{42.15} & \textbf{26.17} \\
    \hline
    SD2 & 1 &  63.00 & 35.63 & 40.32  & 23.57  \\
    SD2 & 5 & \underline{63.50} & \underline{38.21} & \underline{42.57} & \underline{24.65} \\
    SD2 & 20 & \textbf{64.96} & \textbf{39.31} & \textbf{43.64} & \textbf{26.04}\\
    \hline
    SDXL & 1 &  56.58 & 33.05 & 34.75  & 17.94  \\
    SDXL & 5 & \underline{58.14} & \underline{34.43} & \underline{36.01} & \underline{18.73} \\
    SDXL & 20 & \textbf{58.40} & \textbf{35.70} & \textbf{36.43} & \textbf{19.28}\\
    \hline
    SD3 & 1 & \underline{54.56} & 38.99 & 37.26 & 27.04 \\
    SD3 & 5 & 54.03 & \underline{39.64} & \underline{37.37} & \textbf{28.05} \\
    SD3 & 20 & \textbf{55.14} & \textbf{40.02} & \textbf{37.76} & \underline{27.52} \\
    \hline
    SD3.5 & 1 & 51.36 & 36.11 & \underline{36.02} & 22.99 \\
    SD3.5 & 5 & \underline{51.55} & \textbf{36.62} & 35.98 & \textbf{23.90} \\
    SD3.5 & 20 & \textbf{52.04} & \underline{36.51} & \textbf{36.44} & \underline{23.77} \\
    \bottomrule
  \end{tabular}
  }
\end{table}

\paragraph{Effect of Sampling Strategy to Compute ELBO}

The definition of ELBO suggests that timesteps should be sampled uniformly for ELBO computation. We investigated whether alternative sampling strategies could yield better results. We present the results of different sampling strategies for ELBO computation in \tablename~\ref{tab:extend_elbo_timestep_pos}. For the ``Small", ``Middle", and ``Large" strategies, we sample timesteps evenly from $[0, 0.2]$, $[0.4, 0.6]$, and $[0.7, 0.9]$, respectively. For the ``Random" strategy, we sample timesteps randomly from $[0,1]$. For the ``Even" strategy, we sample timesteps evenly from $[0,1]$. We use 10 timesteps for all sampling strategies to ensure a fair comparison. The results align with our theoretical predictions. The ``Random" strategy is the best for computing alignment score, followed by the ``Even" strategy. Since these two strategies yield similar results, we adopt the ``Even" strategy as default to reduce the randomness of our method. Comparing the results of the ``Small", ``Middle", and ``Large" strategies reveals that larger timesteps experimentally improve ELBO computation, which may be related to the noise schedule $\lambda(t)$.

\begin{table}[t]
\caption{ELBO-T2IAlign results on four benchmark datasets when using $\text{ELBO Timesteps}=10$ with different sampling strategies to compute ELBO.}
  \label{tab:extend_elbo_timestep_pos}
  \centering
   \resizebox{\linewidth}{!}{
  \begin{tabular}{cc|cccc}
    \toprule
    \multirow{2}{*}{Model} &\multirow{2}{*}{Strategy}& \multicolumn{4}{c}{\textbf{mIoU} $\uparrow$ }\\
         &  & VOC & Context & COCO & Ade20K \\
    \midrule
    SD1.5 & Small & 67.07 & 40.93 & 39.55 & 23.27 \\
    SD1.5 & Middle & 68.16 & 42.68 & 41.34 & 25.84 \\
    SD1.5 & Large & \textbf{68.80} & 42.97 & 41.46 & 25.79 \\
    SD1.5 & Random & \underline{68.34} & \textbf{43.21} & \textbf{41.86} & \textbf{26.15} \\
    SD1.5 & Even & 68.18 & \underline{43.21} & \underline{41.62} & \underline{26.08} \\
    \hline
    SD2 & Small & 62.67 & 34.69 & 39.93 & 22.25 \\
    SD2 & Middle & 62.57 & 35.54 & 41.48 & 25.06 \\
    SD2 & Large & 62.74 & 36.94 & 41.91 & 25.59 \\
    SD2 & Random & \textbf{63.54} & \underline{37.18} & \textbf{42.61} & \textbf{25.91} \\
    SD2 & Even & \underline{63.38} & \textbf{37.35} & \underline{42.10} & \underline{25.61} \\
    \hline
    SDXL & Small & 51.70 & 19.70 & 24.87 & 8.78 \\
    SDXL & Middle & 52.19 & 20.21 & 25.40 & \textbf{{9.97}} \\
    SDXL & Large & 52.76 & 20.37 & 25.33 & 9.85 \\
    SDXL & Random & \underline{52.77} & \textbf{21.09} & \underline{25.43} & \underline{9.96} \\
    SDXL & Even & \textbf{52.87} & \underline{20.77} & \textbf{25.63} & 9.88 \\
    \hline
    SD3 & Small & 53.36 & 38.39 & 37.61 & 26.14 \\
    SD3 & Middle & \underline{54.57} & 39.86 & \textbf{38.17} & 26.90 \\
    SD3 & Large & 54.55 & \textbf{40.27} & 37.52 & \textbf{28.47} \\
    SD3 & Random & 54.53 & 39.98 & \underline{37.59} & 26.68 \\
    SD3 & Even & \textbf{54.58} & \underline{39.99} & 37.47 & \underline{27.71} \\
    \bottomrule
  \end{tabular}
  }
\end{table}

\paragraph{Evaluations of Different $\gamma$}

\begin{figure}[t]
\centering
\includegraphics[width=1\linewidth]{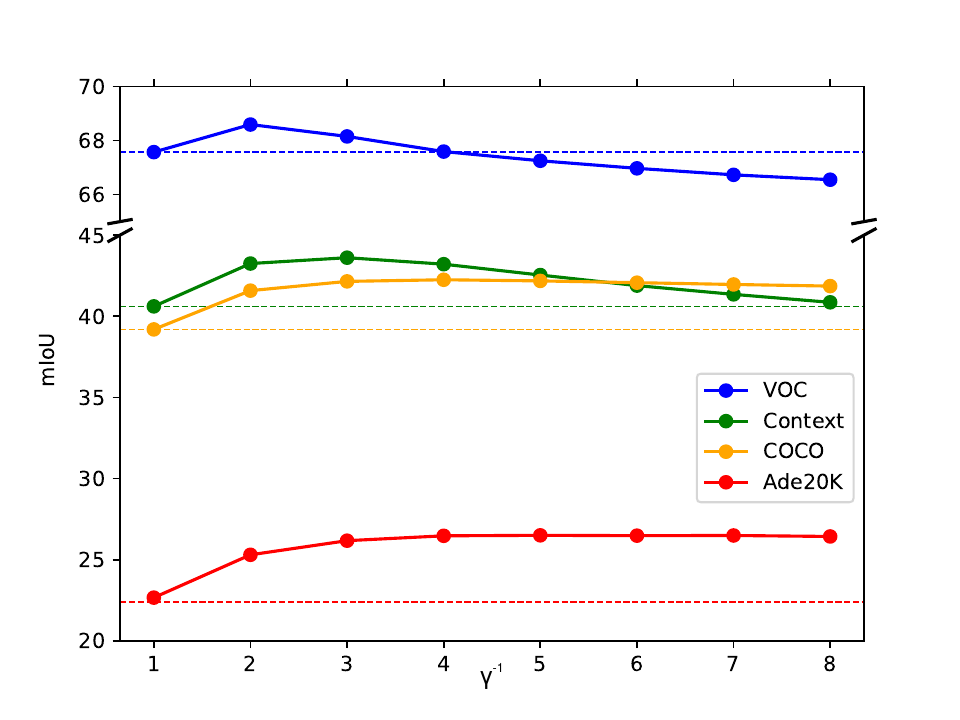}
\caption{\textbf{How $\gamma$ in (\ref{eq:edp_core}) affects the segmentation results using Algorithm~\ref{alg:core}}. The dotted lines indicate $\gamma=1$, which means ELBO calibration is inactive.}
\label{fig:temperature}
\end{figure}

In our method, $\gamma$ in (\ref{eq:edp_core}) is a critical hyperparameter that directly affects the ability of the ELBO to calibrate pixel-text alignment. An inappropriate value of $\gamma$ can even worsen pixel-text alignment. Therefore, we varied $\gamma^{-1} \in \{1,2,3,4,5,6,7,8\}$ to observe its impact on mIoU. Our results show that the performance generally improves as $\gamma$ decreases from $1$ to around $1/2$ or $1/3$, but further decreases in $\gamma$ tend to either plateau or result in a slight decline. This suggests that $\gamma=1/2$ or $\gamma=1/3$ will be the best choice. In our main experiment, we choose $\gamma=1/3$ as the default value. We show the pixel-text alignment results on four benchmark datasets in \figurename~\ref{fig:temperature}. To demonstrate that the effect is not due to the sqrt function itself, we fix alignment score $\{S_i\}_{i=1}^N$ on line 7 of Algorithm~\ref{alg:core}. We set alignment score of each class to the same constant. As \tablename~\ref{tab:fix_temp} shows, the results are better when ELBO is in effect.

\begin{table}[t]
\caption{{ELBO-T2IAlign results on four benchmark datasets when alignment score $S$ is fixed to constant}. The results are better when ELBO is in effect.}
  \label{tab:fix_temp}
  \centering
    \resizebox{\linewidth}{!}{
      \begin{tabular}{cc|cccc}
        \toprule
        \multirow{2}{*}{Method} &\multirow{2}{*}{\begin{tabular}{c} Alignment \\ Score $S$\end{tabular}}& \multicolumn{4}{c}{\textbf{mIoU} $\uparrow$ }\\
             &  & VOC & Context & COCO & Ade20K \\
        \midrule
        SD1.5 & $S=\bm{1/2}$ & 66.61 & 40.29 & 40.13 & 22.90 \\
        SD1.5 & $S=\bm{1/3}$ & 65.11 & 39.72 & 40.09 & 23.04 \\
        SD1.5 & $S=\bm{1/4}$ &  63.28 & 38.69 & 39.58 & 22.87 \\
          SD1.5 & ELB based $S$ & \textbf{68.15} & \textbf{43.61} & \textbf{42.15} & \textbf{26.17} \\
          \hline
          SD2 & $S=\bm{1/2}$ & 61.09 & 35.50 & 39.43 & 21.83 \\
        SD2 & $S=\bm{1/3}$ & 58.45 & 33.22 & 38.97 & 21.24 \\
        SD2 & $S=\bm{1/4}$ &  55.64 & 31.09 & 38.32 & 20.38 \\
          SD2 & ELB based $S$ & \textbf{64.96} & \textbf{39.31} & \textbf{43.64} & \textbf{26.04}\\
          \hline
          SDXL & $S=\bm{1/2}$ & 57.21 & 33.05 & 35.70 & 18.30 \\
        SDXL & $S=\bm{1/3}$ & 55.99 & 32.78 & 35.82 & 18.55 \\
        SDXL & $S=\bm{1/4}$ &  54.39 & 32.49 & 35.58 & 18.75 \\
          SDXL & ELB based $S$ & \textbf{58.40} & \textbf{35.70} & \textbf{36.43} & \textbf{19.28}\\
        \bottomrule
      \end{tabular}
  }
\end{table}

\paragraph{Effect of Different Attention Collecting Timesteps}

\begin{table}[t]
\caption{ELBO-T2IAlign results on four benchmark datasets with different attention collecting steps and strategies.}
  \label{tab:collect_timestep}
  \centering
  \resizebox{\linewidth}{!}{
  \begin{tabular}{ccc|cccc}
    \toprule
    \multirow{2}{*}{Model} &\multirow{2}{*}{Steps}&\multirow{2}{*}{Strategy}& \multicolumn{4}{c}{\textbf{mIoU} $\uparrow$ }\\
         & & & VOC & Context & COCO & Ade20K \\
    \midrule
    SD1.5 & 1 & Small & \underline{67.97} & \underline{42.38} & \underline{41.41} & \underline{25.92} \\
    SD1.5 & 10 & Small & \textbf{68.18} & \textbf{43.21} & \textbf{41.62} & \textbf{26.08} \\
    SD1.5 & 10 & Middle & 57.77 & 40.61 & 30.78 & 22.42 \\
    SD1.5 & 10 & Large & 37.69 & 27.47 & 18.72 & 14.27 \\
    SD1.5 & 10 & Random & 51.95 & 35.95 & 27.17 & 19.62 \\
    \hline
    SD2 & 1 & Small & 62.52 & \underline{39.56} & 37.37 & 23.27 \\
    SD2 & 10 & Small & \textbf{64.62} & 39.06 & \textbf{43.09} & \textbf{25.43} \\
    SD2 & 10 & Middle & \underline{63.50} & \textbf{40.29} & \underline{37.38} & \underline{23.51} \\
    SD2 & 10 & Large & 44.25 & 31.32 & 23.43 & 16.89 \\
    SD2 & 10 & Random & 56.49 & 37.65 & 32.13 & 21.94 \\
    \hline
    SD3 & 1 & Small & \textbf{54.60} & 39.04 & \textbf{37.62} & 26.14 \\
    SD3 & 10 & Small & \underline{54.55} & 39.98 & \underline{37.47} & 26.90 \\
    SD3 & 10 & Middle & 53.38 & \textbf{42.49} & 35.00 & \textbf{30.17} \\
    SD3 & 10 & Large & 39.32 & 32.37 & 22.54 & 20.42 \\
    SD3 & 10 & Random & 46.41 & 38.14 & 29.46 & 26.18 \\
    SD3 & 10 & Even & 49.43 & \underline{40.13} & 32.81 & \underline{27.68} \\
    \bottomrule
  \end{tabular}
  }
\end{table}

For simplicity, we only select one timestep $t$ in Algorithm~\ref{alg:core} to demonstrate our calibration process. In practice, we select multiple timesteps to generate pixel-text alignment results and average them. Different timesteps represent different stages in the diffusion process, with each timestep value affecting the data generated during inference. Both cross-attention and self-attention maps—which are fundamental to obtaining the final mask—are significantly influenced by timesteps, which in turn affect pixel-text alignment. To examine how timesteps influence pixel-text alignment during mask generation, we experiment with different timestep values while maintaining all other hyperparameters constant. To obtain attention maps, we sample timesteps evenly within specific ranges. The parameter ``Steps" in Table~\ref{tab:collect_timestep} indicates the number of sampling timesteps and the ``Strategy" defines the specific sampling range: ``Small" corresponds to $[0,0.2]$, ``Middle" corresponds to $[0.4,0.6]$, ``Large" corresponds to $[0.7,0.9]$, and ``Random" means random sampling from the range $[0,1]$. Table~\ref{tab:collect_timestep} demonstrates that the sampling range of timesteps significantly affects pixel-text alignment. Larger timesteps introduce more noise, which disrupts the original image information and leads to poorer alignment performance. Small timesteps barely alter the visual information contained in the image. Based on these findings, we set Steps=$10$ and use the range $[0,0.2]$ as our default values in the main experiment.

\paragraph{Analysis of Class Imbalance Effects}
To assess the impact of training data distribution, we estimate the occurrence probability of object categories by counting their frequencies within a random 16-million sample of the LAION dataset~\cite{schuhmann2021laion} via regular expression matching. Unlike VOC or Context, COCO and Ade20K exhibit significant long-tail distributions, making them suitable for analyzing data bias. We stratify categories into Rare, Common, and Frequent groups based on frequency tertiles (33rd and 66th percentiles).
As shown in Table~\ref{tab:various_frequency}, the baseline model (SD1.5 without ELB) exhibits a significant performance disparity between ``Common" and ``Rare" categories. This confirms that data bias leads to text-image misalignment, particularly for underrepresented objects. After applying ELBO-T2IAlign, we observe consistent improvements across all frequency groups on both benchmarks, showing that the proposed calibration is effective in most cases where the frozen model still provides usable but biased alignment responses. This indicates that our method effectively mitigates the text-image misalignment stemming from data bias.

\begin{table}[t]
\caption{Comparison results of \textbf{before and after calibration} using our ELBO-T2IAlign (ELB for short) on SD1.5 across objects with different occurrence frequencies on benchmark datasets.}
  \label{tab:various_frequency}
  \centering
  \resizebox{\linewidth}{!}{
    \begin{tabular}{cc|ccc}
      \toprule
      \multirow{2}{*}{Dataset} &\multirow{2}{*}{ELB}& \multicolumn{3}{c}{\textbf{mIoU} $\uparrow$ }\\
           &  & Rare & Common & Frequent \\
      \midrule
      \multirow{2}{*}{COCO} &  & 31.39 & 46.09 & 40.34 \\
        &$\checkmark$ & 33.63 & 50.25 & 43.07 \\
      \hline
      \multirow{2}{*}{Ade20k} &  & 18.93 & 23.15 & 25.60 \\
        &$\checkmark$ & 23.64 & 27.62 & 28.47 \\
      \bottomrule
    \end{tabular}
  }
\end{table}

\subsection{Limitation}
\begin{figure}[!ht]
    \centering
    \includegraphics[width=0.8\linewidth]{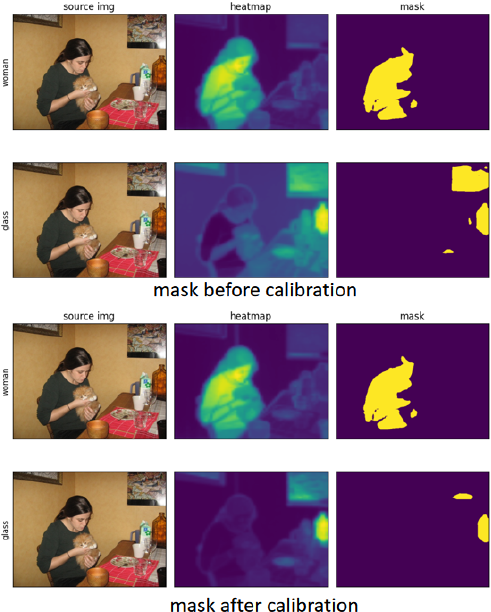}
    \caption{\textbf{Failure case due to missing intrinsic activations.} While the ``woman" is correctly segmented due to the presence of valid cross-attention signals, the ``glass" on the table fails to trigger any activation in the frozen diffusion model.}
    \label{fig:woman_glass}
\end{figure}

Most evaluated cases fall into the common setting where the frozen model provides weak or biased but usable attention responses, which ELBO-T2IAlign can calibrate effectively. While our ELBO-T2IAlign demonstrates effectiveness in calibrating pixel-text alignment, our method relies on the intrinsic recognition capability of the frozen diffusion model. As a calibration approach, it re-weights existing cross-attention signals rather than generating new ones. Consequently, if the base model completely fails to activate the attention map for a specific object (e.g., failing to recognize ``glass", resulting in zero activation, information mix-up~\cite{chen2024cat}), our method cannot recover the missing mask. A concrete example is illustrated in \figurename~\ref{fig:woman_glass}. The model successfully segments the ``woman" based on valid attention signals. Conversely, the ``glass" on the table lacks any initial activation, resulting in a failure to generate a correct mask in both the baseline and our calibrated results. Exploring this limitation represents an interesting direction for future work.

\section{Conclusions}
\label{sec:conclusion}

Our research introduces a novel training-free and generic approach called ELBO-T2IAlign, which leverages the evidence lower bound (ELBO) of likelihood to correct pixel-text misalignment in pre-trained diffusion models. We evaluated the alignment capabilities of popular diffusion models and uncovered the training data biases that lead to misalignment, particularly in images with small, occluded, or rare class objects. Our proposed method offers an effective alignment calibration solution that is agnostic to the underlying cause of misalignment and compatible with diverse diffusion model architectures. The results indicate that ELBO-T2IAlign calibrates a shared pixel-text alignment signal rather than targeting only a single downstream behavior. Its training-free and architecture-agnostic nature makes it practically applicable to different frozen diffusion backbones without additional annotations, retraining, or model modifications. This work paves the way for more effective downstream tasks in image segmentation, image editing, and controllable generation. This work relies on the accuracy of the ELBO likelihood estimated by pre-trained diffusion models. Nevertheless, despite this imperfect estimation, our method consistently improves performance.

\section*{Acknowledgments}
This work was supported in part by National Natural Science Foundation of China (No.62461160331, No.62132001, No.62572039), in part by Huawei-BUAA Joint Lab, in part by the NSFC/RGC Collaborative Research Scheme (CRS\_HKU703/24). Dr. Xu's research work described in this paper was conducted in the JC STEM Lab of Multimedia and Machine Learning funded by The Hong Kong Jockey Club Charities Trust.
 
%
\bibliographystyle{IEEEtran}
\bibliography{IEEEabrv,main}

\end{document}


\title{ELBO-T2IAlign: A Generic ELBO-Based Method for Calibrating Pixel-level Text-Image Alignment in Diffusion Models}

\author{Supplementary Material}



\maketitle

In the supplementary material, we report derivations of ELBO for additional diffusion loss functions, more implementation details, additional experimental results and analyses, as well as more visualizations.

\section{Derivation of ELBO Computation}
\label{sec:elbo_derivation}
As mentioned in the main paper, different diffusion training objectives can be derived as weighted ELBO objectives. In this section, we derive ELBO weights in the formulations of different diffusion loss functions, including score matching objective, conditional flow matching objective, noise prediction objective, velocity prediction objective, and sample prediction objective. We evaluate our method under noise prediction objective (Stable Diffusion v1 series), velocity prediction objective (Stable Diffusion v2 series), and conditional flow matching objective (Stable Diffusion 3.5), leaving other objectives to future work.

\paragraph{Score Matching Objective}

Score matching objective~\cite{songscore} uses a neural network to approximate score function $\nabla_{z_t}\text{ln}q(z_{t}|x)$. The diffusion forward process can be modeled as an SDE:
\begin{equation}
    \mathrm{d}z_t=f(z_t,t)\mathrm{d}t+g(t)\mathrm{d}\mathrm{w},
\end{equation}
where $\mathrm{w}$ is standard Brownian motion, $f(\cdot,t):\mathbb{R}^{d}\xrightarrow{}\mathbb{R}^{d}$ is a vector-valued function called drift coefficient, and $g(\cdot):\mathbb{R}\xrightarrow{}\mathbb{R}$ is a scalar function called diffusion coefficient. Using $\nabla_{z_t}\text{ln}q(z_{t}|x)$, we can sample $x$ by reverse-time SDE:
\begin{equation}
    \mathrm{d}x=[f(z_t,t)-g(t)^2\nabla_{z_t}\text{ln}q(z_{t}|x)]\mathrm{d}t+g(t)\mathrm{d}\mathrm{w}.
\end{equation}
Score matching objective is given by:
\begin{multline}
\mathcal{L}^{score}_\lambda(x,c)\\=\frac{1}{2}\mathbb{E}_{t\sim\mathcal{U}(0,1),\epsilon\sim\mathcal{N}(0,1)}[\left\|s_\theta(z_{t},t,c)-\nabla_{z_t}\text{ln}q(z_{t}|x)\right\|_2^2],
\end{multline}
where $s_\theta(z_{t},t,c)$ is a neural network. By definition of the forward process, we have:
\begin{align}
    q(z_t|x,c)&=\mathcal{N}(z_t;\alpha_tx,\sigma_t^{2}I)\\
    &=\frac{1}{\sqrt{(2\pi)^k|\Sigma|}}e^{-\frac{1}{2\sigma_t^2}(z_t-\alpha_tx)^{T}(z_t-\alpha_tx)}
\end{align}
We can compute $\nabla_{z_t}\text{ln}q(z_{t}|x)$ by:
\begin{align}
    \nabla_{z_t}\text{ln}q(z_{t}|x)&=\nabla_{z_t}(-\text{ln}\sqrt{(2\pi)^k|\Sigma|}-\frac{1}{2\sigma_t^2}\left\|z_t-\alpha_tx\right\|_2^2) \\
    &=-\frac{1}{\sigma_t^2}(z_t-\alpha_tx)\\
    &=-\frac{\epsilon}{\sigma_t}
\end{align}
Therefore, the score matching objective can be rewritten as:
\begin{equation}
    \mathcal{L}^{score}_\lambda(x,c)=\frac{1}{2}\mathbb{E}_{t\sim\mathcal{U}(0,1),\epsilon\sim\mathcal{N}(0,1)}[\frac{1}{\sigma_t^2}\left\|\epsilon_\theta(z_{t},t,c)-\epsilon\right\|_2^2].
\end{equation}
Considering that ELBO objective is:
\begin{equation}
  ELB_\lambda(x,c)=\frac{1}{2}\mathbb{E}_{t\sim\mathcal{U}(0,1),\epsilon\sim\mathcal{N}(0,1)}[-\frac{\mathrm{d}\lambda}{\mathrm{d}t}\cdot\left\|\epsilon_\theta(z_{t},t,c)-\epsilon\right\|_2^2].
\end{equation}
We can derive $\omega^{score}(t)=-1/({\sigma_t^{2}}\cdot\lambda'(t))$.

\paragraph{Conditional Flow Matching Objective}

Conditional flow matching objective~\cite{lipman_flow_2022} models the diffusion process as an ODE:
\begin{equation}
    \mathrm{d}z_t=u_t(z_t|\epsilon)\mathrm{d}t,
\end{equation}
where $u_t(z_t|\epsilon)$ is a conditional vector field that generates conditional probability path $q(z_t|\epsilon)$. We can reverse the diffusion forward process using a neural network $\mathrm{v}_\theta(z_t,t,c)$ to estimate $u_t(z_t|\epsilon)$ and solve the reverse ODE\@. The conditional flow matching objective is given by:
\begin{equation}
\mathcal{L}^{flow}_\lambda(x,c)=\frac{1}{2}\mathbb{E}_{t\sim\mathcal{U}(0,1),\epsilon\sim\mathcal{N}(0,1)}[\left\|\mathrm{v}_\theta(z_{t},t,c)-u_t(z_t|\epsilon)\right\|_2^2].
\end{equation}
Conditional vector field $u_t(z_t|\epsilon)$ is given by:
\begin{align}
    u_t(z_t|\epsilon)&=\frac{\mathrm{d}z_t}{\mathrm{d}t}\\
    &=\frac{\mathrm{d}(\alpha_tx+\sigma_t\epsilon)}{\mathrm{d}t}\\
    &=\alpha_t'x+\sigma_t'\epsilon\\
    &=\alpha_t'(\frac{z_t-\sigma_t\epsilon}{\alpha_t})+\sigma_t'\epsilon\\
    &=\frac{\alpha_t'}{\alpha_t}z_t-\sigma_t(\frac{\alpha_t'}{\alpha_t}-\frac{\sigma_t'}{\sigma_t})\epsilon\\
    &=\frac{\alpha_t'}{\alpha_t}z_t-\frac{\sigma_t}{2}\lambda'(t)\epsilon
\end{align}
Therefore, conditional flow matching objective can be rewritten as:
\begin{equation}
    \mathcal{L}^{flow}_\lambda(x,c)=\frac{1}{2}\mathbb{E}_{t\sim\mathcal{U}(0,1),\epsilon\sim\mathcal{N}(0,1)}[\frac{\sigma_t^{2}\lambda'(t)^2}{4}\left\|\epsilon_\theta(z_{t},t,c)-\epsilon\right\|_2^2].
\end{equation}
Comparing $\mathcal{L}^{flow}_\lambda(x,c)$ with $ELB_\lambda(x,c)$, we can derive $\omega^{flow}(t)=-(\sigma_t^{2}\cdot\lambda'(t))/4$.

\paragraph{Noise Prediction Objective}

Noise prediction objective~\cite{ho_denoising_2020} is very similar to ELBO objective. Noise prediction objective suggests directly using a neural network $\epsilon_\theta(z_t,t,c)$ to predict ground-truth noise $\epsilon$ and omitting $\lambda'(t)$:
\begin{equation}  \mathcal{L}^{noise}_\lambda(x,c)=\frac{1}{2}\mathbb{E}_{t\sim\mathcal{U}(0,1),\epsilon\sim\mathcal{N}(0,1)}[\left\|\epsilon_\theta(z_{t},t,c)-\epsilon\right\|_2^2].
\end{equation}
So we can also directly infer $\omega^{noise}(t)=-1/\lambda'(t)$.

\paragraph{Velocity Prediction Objective}

Velocity prediction~\cite{salimans_progressive_2021} is based on the DDIM~\cite{song_denoising_2020} update rule. Specifically, if we define $\phi_t=\arctan(\sigma_t/\alpha_t)$, then latent $z$ can be expressed as:
\begin{equation}
    z_\phi=\cos(\phi)x+\sin(\phi)\epsilon.
\end{equation}
The velocity of $z_\phi$ is defined as the derivative of $z_\phi$ with respect to $\phi$:
\begin{equation}
    v_\phi=\frac{\mathrm{d}z_\phi}{\mathrm{d}\phi}=\cos(\phi)\epsilon-\sin(\phi)x.
\end{equation}
With the definition of velocity $v_\phi$, sample $x$ and noise $\epsilon$ can be rewritten in terms of $v_\phi$ and $z_\phi$:
\begin{align}
    x&=\cos(\phi)z_\phi-\sin(\phi)v_\phi\\
    \epsilon&=\sin(\phi)z_\phi+\cos(\phi)v_\phi
\end{align}
Intriguingly, the DDIM update rule can also be rewritten as:
\begin{equation}
    z_{\phi_{s}}=\cos(\phi_s-\phi_t)z_{\phi_t}+\sin(\phi_s-\phi_t)v_{\phi_t}.
\end{equation}
From this perspective, DDIM thus evolves $z_\phi$ by moving it on a circle in the $(z_{\phi_t},v_{\phi_t})$ basis, along the $-v_{\phi_t}$ direction. The velocity prediction objective is given by:
\begin{equation}
    \mathcal{L}^{v}_\lambda(x,c)=\frac{1}{2}\mathbb{E}_{t\sim\mathcal{U}(0,1),\epsilon\sim\mathcal{N}(0,1)}[\left\|v_\theta(z_{t},t,c)-v_t\right\|_2^2],
\end{equation}
where $v_\theta(z_{t},t,c)$ is a neural network, and $v_t=\alpha_t\epsilon-\sigma_tx$. We can rewrite $v_t$ in terms of $z_t$ and $\epsilon$:
\begin{align}
    v_t&=\alpha_t\epsilon-\sigma_tx\\
    &=\alpha_t\epsilon-\sigma_t(\frac{z_t-\sigma_t\epsilon}{\alpha_t})\\
    &=-\frac{\sigma_t}{\alpha_t}z_t+\frac{\alpha_t^2+\sigma_t^2}{\alpha_t}\epsilon
\end{align}
Hence, we can rewrite the velocity prediction objective:
\begin{equation}
    \mathcal{L}^{v}_\lambda(x,c)=\frac{1}{2}\mathbb{E}_{t\sim\mathcal{U}(0,1),\epsilon\sim\mathcal{N}(0,1)}[\frac{(\alpha_t^2+\sigma_t^2)^2}{\alpha_t^2}\left\|\epsilon_\theta(z_{t},t,c)-\epsilon\right\|_2^2].
\end{equation}
Finally, we can infer $\omega^v(t)=-(\alpha^2_{t}+\sigma^2_{t})^2/(\alpha^2_{t}\cdot\lambda'(t))$.

\paragraph{Sample Prediction Objective}

We can also directly predict sample $x$ as an objective~\cite{sohl-dickstein_deep_2015}:

\begin{equation}
    \mathcal{L}^{sample}_\lambda(x,c)=\frac{1}{2}\mathbb{E}_{t\sim\mathcal{U}(0,1),\epsilon\sim\mathcal{N}(0,1)}[\left\|x_\theta(z_{t},t,c)-x\right\|_2^2].
\end{equation}

Recalling that $x=\frac{z_t-\sigma_t\epsilon}{\alpha_t}$, we can rewrite the sample prediction objective as:
\begin{equation}
    \mathcal{L}^{sample}_\lambda(x,c)=\frac{1}{2}\mathbb{E}_{t\sim\mathcal{U}(0,1),\epsilon\sim\mathcal{N}(0,1)}[\frac{\sigma_t^2}{\alpha_t^2}\left\|\epsilon_\theta(z_{t},t,c)-\epsilon\right\|_2^2].
\end{equation}
Thus, we have $\omega^{sample}(t)=-1/(\lambda'(t)\cdot e^{\lambda(t)})$.

\section{More Implementation Details}
\label{sec:more_implementation_detail}
In this section, we provide more implementation details of our proposed ELBO-T2IAlign and other baseline methods, including DiffSegmenter, DAAM, OVAM, Semantic DiffSeg, and DiffPNG.
\paragraph{ELBO-T2IAlign} For a given image, we first resize it to the maximum input size accepted by the diffusion model. For example, we resize images to $(512,512)$ for Stable Diffusion v1.5 and to $(768,768)$ for Stable Diffusion v2.1. Second, we randomly generate noise $\epsilon$ to compute ELBO and collect attention maps. To compute ELBO, we run diffusion on the given image using generated $\epsilon$, and compute ELBO using the above formulation. Following training practice of diffusion models, we always use all text encoders to produce text embeddings. To collect attention maps, we also run diffusion on the given image, and use a prompt that contains all class names to predict $\epsilon$. We format the prompt for collecting attention maps as ``a photo of class\_A, class\_B,\ldots". We can record the position of each class in the prompt to extract cross-attention maps. If a class name contains multiple words, we simply average the cross-attention maps with respect to these words. We can also extract self-attention maps when collecting cross-attention maps. Finally, we aggregate attention maps and generate final masks. We aggregate cross-attention maps by averaging all attention heads and assigning high weights to low-resolution cross-attention maps. For example, Stable Diffusion v1.5 has cross-attention maps at 4 different resolutions, and we take a weighted average of these cross-attention maps, assigning larger weights to lower-resolution cross-attention and smaller weights to higher-resolution cross-attention. We aggregate self-attention maps by averaging all attention heads and assigning high weights to high-resolution self-attention maps. Taking Stable Diffusion v1.5 as an example, we only average the highest-resolution self-attention maps and discard others. Our attention aggregation strategy is based on previous work~\cite{wang_diffusion_2024,tian_diffuse_2024}. After using alignment score $S$ to refine cross-attention maps, we simply multiply self-attention maps and refined cross-attention maps to get final masks.

\paragraph{DiffSegmenter~\cite{wang_diffusion_2024}} DiffSegmenter pipeline is very similar to ours. The original DiffSegmenter uses BLIP to enhance text prompt. For fair comparison, we do not use BLIP for DiffSegmenter pipeline. However, ELBO-T2IAlign still surpasses DiffSegmenter with BLIP\@. We follow the original protocol and code of DiffSegmenter to generate masks. For an image that has only one class, we generate a prompt like ``a photo of (class\_A)++" to collect and aggregate attention maps. Here, the sign ``++" indicates prompt reweighting in the original paper. For an image with multiple classes, we generate multiple prompts and obtain final masks in the order of class names. To improve DiffSegmenter with alignment score, we simply refine cross-attention maps with our formulation and keep other operations the same.

\paragraph{DAAM~\cite{tang_what_2023}}
DAAM is based on cross-attention mechanisms. In the original method, cross-attention-related data are obtained during the inference process of generating images. Since our task is based on real images, we iteratively add random noise to the given image and perform denoising with the image and a prompt. We generate a prompt like ``a photo of class\_A, class\_B,\ldots". Then, we collect all heatmaps and compute every category's heatmap according to the original pipeline. Finally, we resize the heatmap to the target size using bilinear interpolation, then apply min-max normalization and scale it to the $[0, 255]$ range. In terms of hyperparameters, we uniformly sample 20 points between $[20, 200]$ as timesteps for iteration, and other parameters are consistent with the default parameters provided in the original method.

\paragraph{OVAM~\cite{marcos-manchon_open-vocabulary_nodate}}
OVAM is similar to DAAM; thus, the way in which we obtain attention-related data, generate prompts, and perform heatmap post-processing is the same as the DAAM method mentioned above. OVAM also includes a token optimization operation. To obtain optimized embedding for every category, we first generate the ``context\_sentence" like ``a photo of class\_A, class\_B,\ldots", then specify ``text" for a specific category to obtain initial embedding. Next, based on ground truth, we extract the corresponding mask for the specific category as ``embedding\_target" and use it to optimize the embedding. The optimized embedding is then saved in a file format. Finally, we use optimized embedding and attention-related data to get the mask. We uniformly sample 20 points between $[20, 200]$ as timesteps, and other hyperparameters are consistent with the default parameters provided in the original method.

\paragraph{Semantic DiffSeg~\cite{tian_diffuse_2024}}
DiffSeg is an unsupervised zero-shot segmentation method. Therefore, even though we have adapted it to our task, there is still a gap. We use its extended version, which is called Semantic DiffSeg. Instead of using BLIP, we generate captions like ``a photo of class\_A, class\_B,\ldots, and background" based on given categories. We extract category indices and feed (index, category) tuples into the model to get the mask. All hyperparameters are consistent with the default parameters provided in the original method.

\paragraph{DiffPNG~\cite{yang_exploring_2024}}
DiffPNG is also based on cross-attention and self-attention mechanisms. DiffPNG originally extracts noun positions from image captions using the CLIP model. To ensure a fair comparison, we generate prompts like ``a photo of class A, class B,\ldots", similar to what was mentioned above. We use the output of the LSP module from DiffPNG as the final heatmap. This module utilizes $16\times16$ resolution cross-attention to obtain high-confidence pixels as anchor points and aggregates the corresponding $32\times32$ resolution self-attention at those anchor point locations to generate an enhanced mask.

\section{Additional Experiments and Analyses}
\label{sec:additional_exp}
This section presents additional experiments and analyses. We first analyze the computational cost. Then, since our ELBO-T2IAlign method is generic and can be applied to other baseline methods, we conduct further experiments applying ELBO-T2IAlign to other baselines. Lastly, to better analyze the hyper-parameters of ELBO-T2IAlign, we also report results with varying sampling timesteps and attention-collection timesteps.

\paragraph{Applying ELBO-T2IAlign to Other Baselines}

We can also use alignment scores from our method to refine cross-attention maps in other methods. DiffSegmenter~\cite{wang_diffusion_2024} combines cross-attention and self-attention maps to generate final masks. We only refine cross-attention and keep other parts the same. For DAAM~\cite{tang_what_2023}, we precompute alignment scores for each category in images and then refine the generated heatmap through the score. The results are shown in \tablename~\ref{tab:extend_sota_result}. The results show that alignment scores from our method also consistently improve the performance of other methods.

\begin{table}[t]
\caption{\textbf{Results of zero-shot referring image segmentation on four benchmark datasets.} We generate final masks using the best threshold for each method, respectively.}
  \label{tab:extend_sota_result}
  \centering
   \resizebox{\linewidth}{!}{
  \begin{tabular}{cc|cccc}
    \toprule
    \multirow{2}{*}{Method} &\multirow{2}{*}{ELB}& \multicolumn{4}{c}{\textbf{mIoU} $\uparrow$ }\\
         &  & VOC & Context & COCO & Ade20K \\
    \midrule
    DAAM& & 48.11 & 26.13 & 29.16 & 15.02 \\
    DAAM&$\checkmark$& \textbf{49.21}& \textbf{27.10} & \textbf{32.05} & \textbf{18.15} \\
    \hline
    DiffSegmenter& & 68.68 & 39.47 & 38.88 & 22.88\\
    DiffSegmenter&$\checkmark$& \textbf{68.87} & \textbf{42.75} & \textbf{42.24} & \textbf{26.02} \\
    \bottomrule
  \end{tabular}
  }
\end{table}

\paragraph{ELBO-T2IAlign Results of More Diffusion Models}

We also evaluated our method on Stable Diffusion v1.2, Stable Diffusion v1.3, and the Stable Diffusion v2-base model. The results, shown in \tablename~\ref{tab:extend_various_model}, further demonstrate the generalizability of our approach across different diffusion models.

\begin{figure}[t]
    \centering
    \includegraphics[width=1\linewidth]{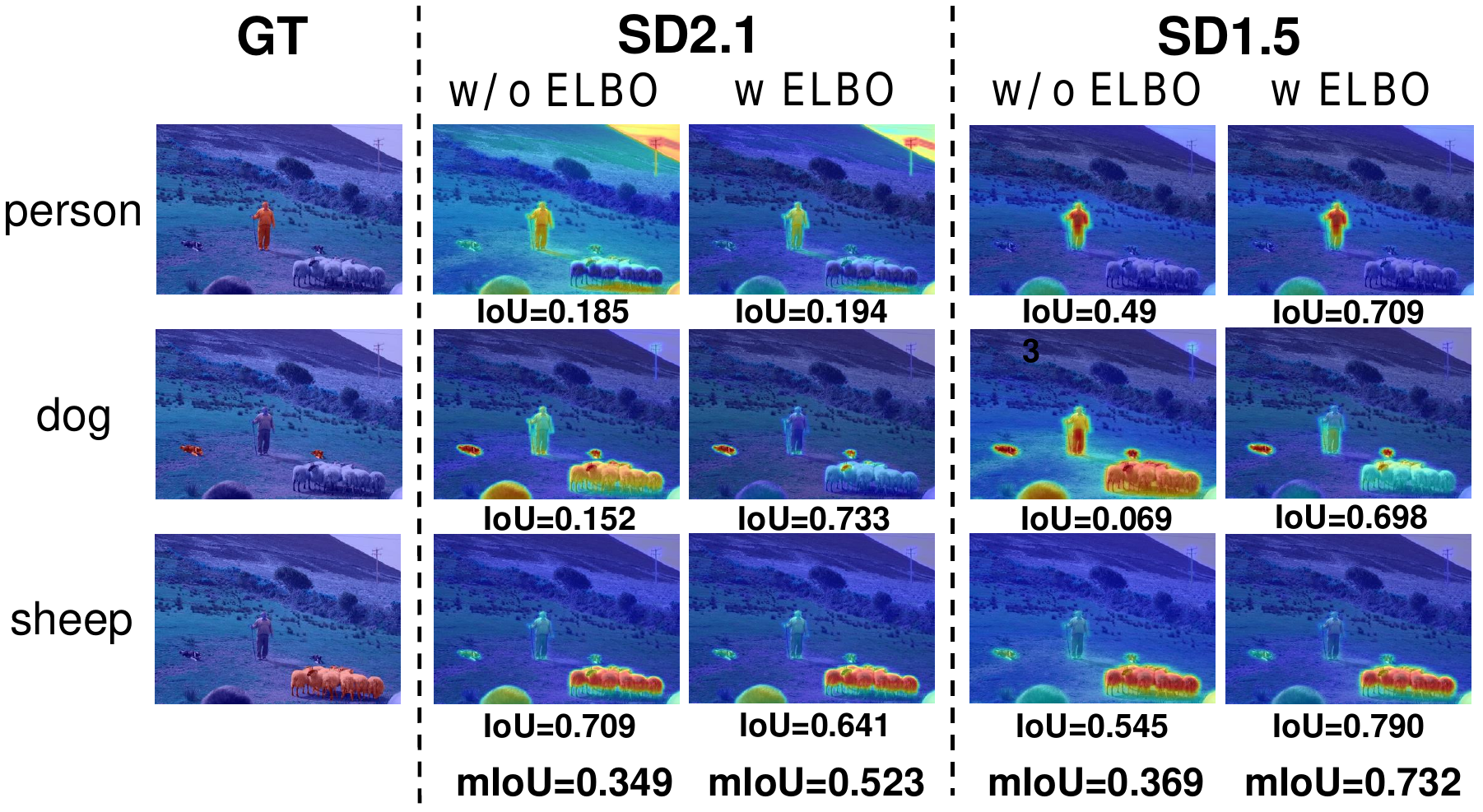}
    \caption{\textbf{Heatmap comparison of the baseline ($\gamma=1$) and our ELBO-T2IAlign ($\gamma=1/3$)}.}
    \label{figure:temp_ablation_visual}
\end{figure}

\begin{table}[t]
  \caption{Comparison results of \textbf{before and after calibration} using our ELBO-T2IAlign (ELB for short) across more diffusion models on four benchmark datasets.}
  \label{tab:extend_various_model}
  \centering
   \resizebox{\linewidth}{!}{
  \begin{tabular}{cc|cccc}
    \toprule
    \multirow{2}{*}{Model} &\multirow{2}{*}{ELB}& \multicolumn{4}{c}{\textbf{mIoU} $\uparrow$ }\\
         &  & VOC & Context & COCO & Ade20K \\
    \midrule
    SD1.2 &  & 68.05 & 40.30 & 38.69 & 22.56 \\
    SD1.2 & $\checkmark$ & \textbf{68.47} & \textbf{43.52} & \textbf{41.78} & \textbf{26.14} \\
    \hline
    SD1.3 &  & 67.78 & 40.87 & 38.59 & 23.25 \\
    SD1.3 & $\checkmark$ & \textbf{68.38} & \textbf{43.79} & \textbf{41.72} & \textbf{26.69} \\
    \hline
    SD2-base &  & 65.85 & 39.40 & 38.94 & 23.11 \\
    SD2-base & $\checkmark$ & \textbf{66.82} & \textbf{42.10} & \textbf{43.42} & \textbf{26.73} \\
    \bottomrule
  \end{tabular}
  }
\end{table}
\figurename~\ref{figure:temp_ablation_visual} visualizes some results of the baseline and the proposed ELBO-T2IAlign. Our method significantly reduces the activation of the class ``dog" in regions of other classes and minimizes confusion between classes.

\paragraph{ELBO-T2IAlign Results with Different Sampling Timesteps}

As mentioned in the main paper, the number of sampling timesteps can affect the alignment score. Both intuitively and theoretically, the computed ELBO becomes more accurate with more sampling steps. We present comprehensive results for different numbers of sampled timesteps in \tablename~\ref{tab:extend_elbo_timestep}. Our experiments show that increasing ELBO timesteps leads to better results across various diffusion architectures. However, to balance accuracy with computational efficiency, we set $20$ steps as our default ELBO timesteps.

\begin{table}[t]
\caption{\textbf{ELBO-T2IAlign results on four benchmark datasets when using different numbers of timesteps to compute ELBO}. ELBO Timesteps denotes the number of sampling timesteps used to compute ELBO\@. We sample timesteps evenly from $[1,999)$.}
  \label{tab:extend_elbo_timestep}
  \centering
   \resizebox{\linewidth}{!}{
  \begin{tabular}{cc|cccc}
    \toprule
    \multirow{2}{*}{Model} &\multirow{2}{*}{Steps}& \multicolumn{4}{c}{\textbf{mIoU} $\uparrow$ }\\
         &  & VOC & Context & COCO & Ade20K \\
    \midrule
    SD1.5 & 1 & 67.35 & 40.58 & 39.80 & 24.58 \\
    SD1.5 & 5 & 68.15 & 42.28 & 41.36 & 24.93 \\
    SD1.5 & 10 & 68.18 & 43.21 & 41.62 & 26.08 \\
    SD1.5 & 20 & 68.15 & 43.61 & 42.15 & 26.17 \\
    SD1.5 & 30 & \textbf{68.70} & \underline{43.83} & \underline{42.26} & \underline{26.62} \\
    SD1.5 & 40 & \underline{68.59} & \textbf{43.95} & \textbf{42.31} & \textbf{26.85} \\
    \hline
    SD2 & 1 &  63.00 & 35.63 & 40.32  & 23.57  \\
    SD2 & 5 & 63.50 & 38.21 & 42.57 & 24.65 \\
    SD2 & 10 & 64.62 & 39.06 & 43.09 & 25.44 \\
    SD2 & 20 & \textbf{64.96} & 39.31 & 43.64 & 26.04\\
    SD2 & 30 & 64.91 & \underline{39.68} & \underline{43.76} & \underline{26.25} \\
    SD2 & 40 & \underline{64.94} & \textbf{39.72} & \textbf{43.79} & \textbf{26.33} \\
    \bottomrule
  \end{tabular}
  }
\end{table}

\section{More Details and Results on Image Generation and Editing}
\label{sec:additional_app}
As mentioned in the main paper, our method opens up possibilities for more downstream tasks beyond image segmentation, particularly image editing and controllable compositional generation. In this section, we demonstrate how ELBO-T2IAlign improves image editing and controllable generation, and report additional results.

\paragraph{Improving Image Editing with ELBO-T2IAlign}

\begin{figure}[t]
    \centering
    \includegraphics[width=1\linewidth]{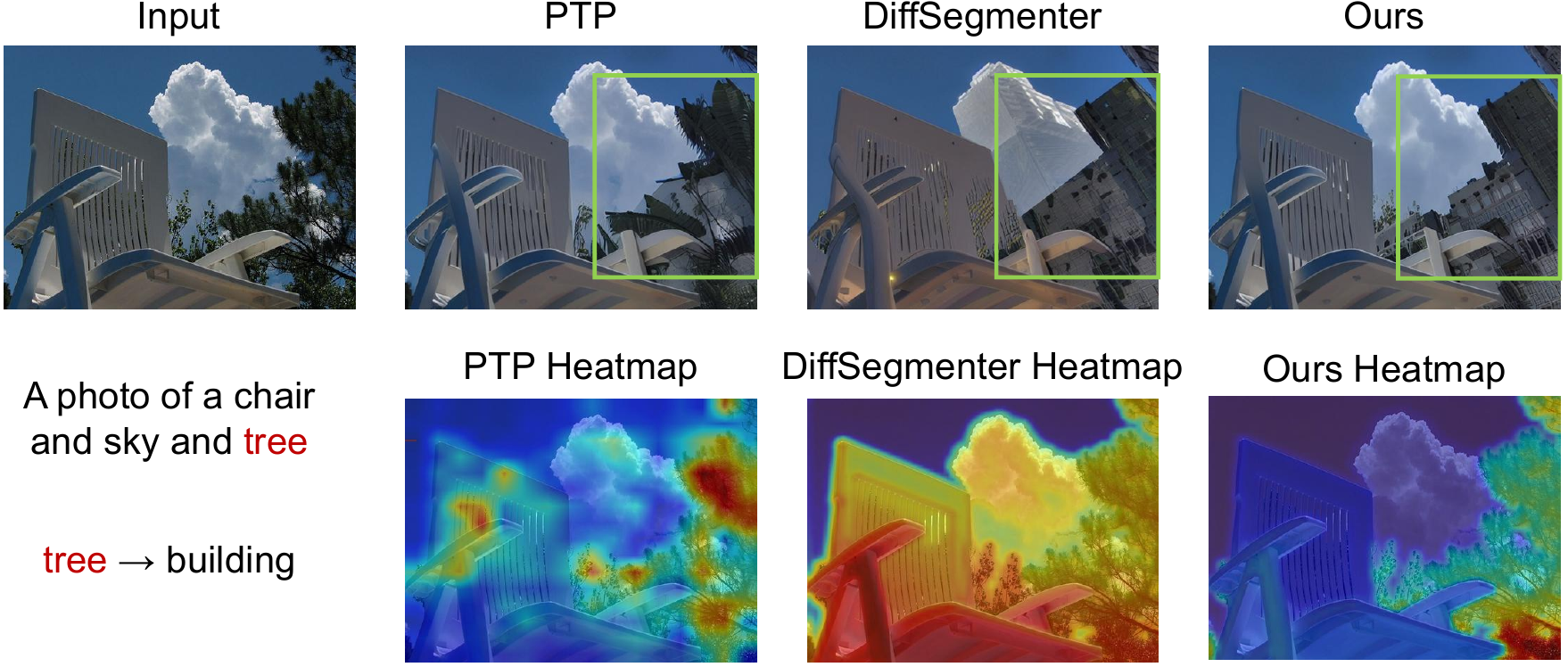}
    \caption{\textbf{Comparison of image editing based on PTP~\cite{hertz_prompt--prompt_2022}} with Null-text Inversion~\cite{mokady_null-text_2023}.}
    \label{figure:edit_compare}
\end{figure}

To demonstrate how our ELBO-T2IAlign method can enhance image editing, we conduct experiments on text-controlled image editing tasks.
Specifically, we use the method of PTP~\cite{hertz_prompt--prompt_2022} with Null-text Inversion~\cite{mokady_null-text_2023} as a real image editing baseline method. In the editing process of PTP, both the original image, which is reconstructed from random noise, and the edited image are produced simultaneously under the Null-Text Inversion constraints. For example, given an image with a chair, sky, and a tree as shown in \figurename~\ref{figure:edit_compare}, if we want to edit ``tree" into ``building", the attention for the target object ``building" is replaced with the attention for ``tree". In the PTP method, precise cross-attention maps are crucial for successful image editing. Both DiffSegmenter~\cite{wang_diffusion_2024} and our ELBO-T2IAlign aim to enhance the cross-attention maps. To evaluate how the enhanced cross-attention maps affect image editing, we replace PTP's cross-attention map of ``building" with the improved heatmaps generated by DiffSegmenter and our ELBO-T2IAlign.
More concretely, we obtain the optimized heatmap, which has a value range of $[0,1]$. We scale the heatmap based on the cross-attention values of ``tree" and replace the cross-attention for the target object ``building" with the scaled heatmap. As shown in \figurename~\ref{figure:edit_compare}, compared to DiffSegmenter and the original PTP method, ELBO-T2IAlign is able to obtain more accurate heatmaps, allowing more precise editing control.

\begin{figure}[t]
    \centering
    \includegraphics[width=1\linewidth]{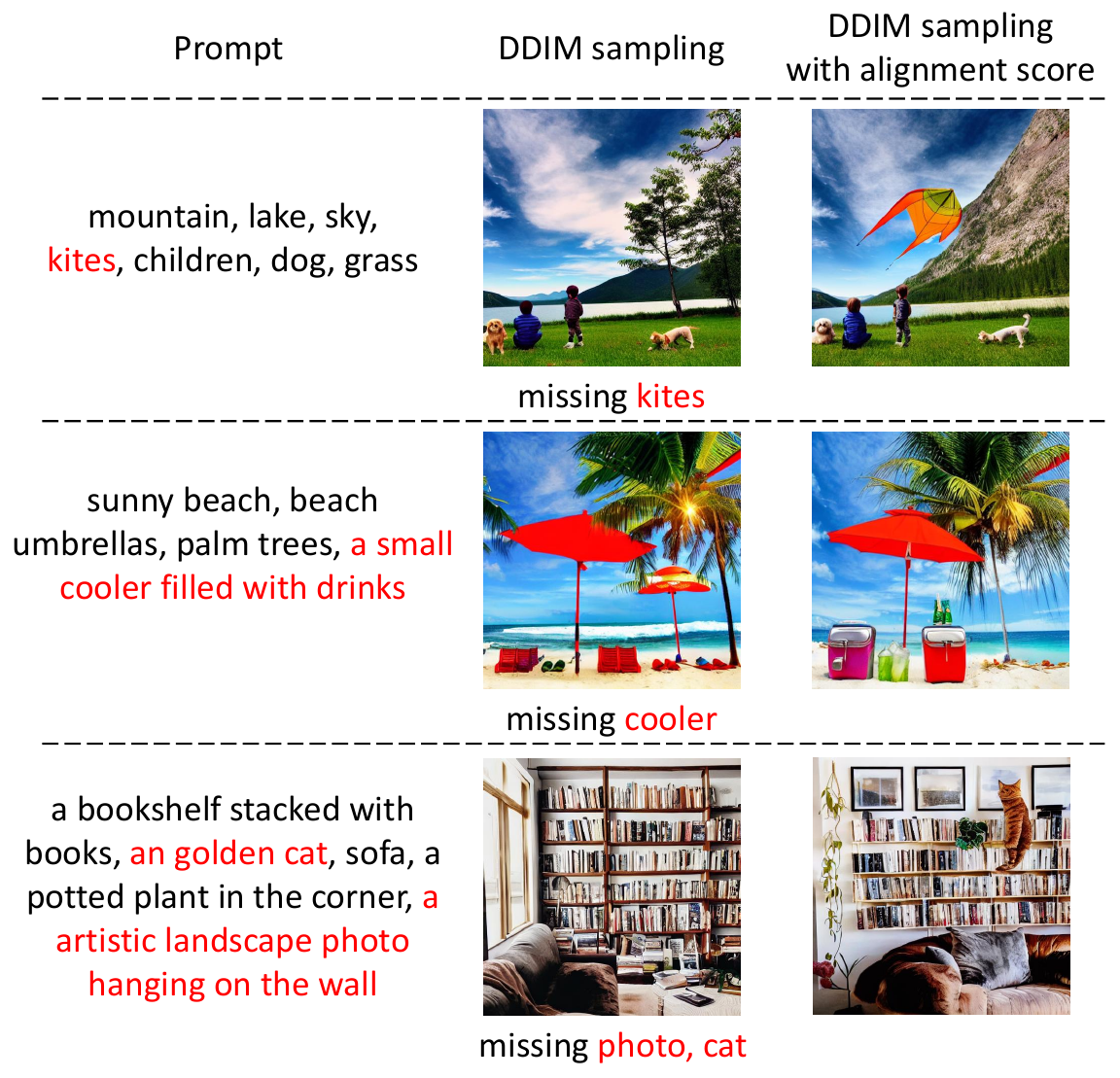}
    \caption{\textbf{Comparison of Stable Diffusion v1.5 compositional image generation based on DDIM with ours.} DDIM sampling often loses some prompt details, while alignment-score-enhanced DDIM captures the prompt more accurately.}
    \label{figure:generate_compare}
\end{figure}

\paragraph{Improving Image Generation with ELBO-T2IAlign}

Our method enhances compositional generation by leveraging the proposed alignment scores. When given an image generation prompt $prompt_{all}$ ``A, B, C,\ldots", diffusion models tend to emphasize words with strong semantic meanings while neglecting finer details~\cite{rassin_linguistic_2023}. Since our alignment score can identify this bias, we combine it with prompt reweighting to improve the quality of compositional generation.
Specifically, we denote a specific class in $prompt_{all}$ and its corresponding alignment score as $prompt_{cls}$ and $S_{cls}$, respectively. When $S_{cls}$ is low, we increase the weight of $prompt_{cls}$.
This method dynamically balances the weights of tokens in $prompt_{all}$ during image sampling, ensuring the generated image aligns better with $prompt_{all}$. We show our results in \figurename~\ref{figure:generate_compare}. Our method demonstrates improved capability to capture the diverse semantics within the prompt.

To compute the alignment score during the sampling process, we assume that the $\epsilon_{all}$ predicted by diffusion models from $prompt_{all}$ is accurate. We denote the prediction generated from $prompt_{cls}$ by diffusion models as $\epsilon_{cls}$. At each sampling step, we first compute a single-step alignment score between $\epsilon_{cls}$ and $\epsilon_{all}$ using the formulation described in the main paper. We then use this alignment score for prompt reweighting to balance the semantics of each $prompt_{cls}$. Finally, we use the reweighted prompt to sample the latent for subsequent timesteps.


\section{More Visualization Results}
\label{sec:more_visual_result}
\begin{figure*}[t]
    \centering
    \includegraphics[width=\linewidth]{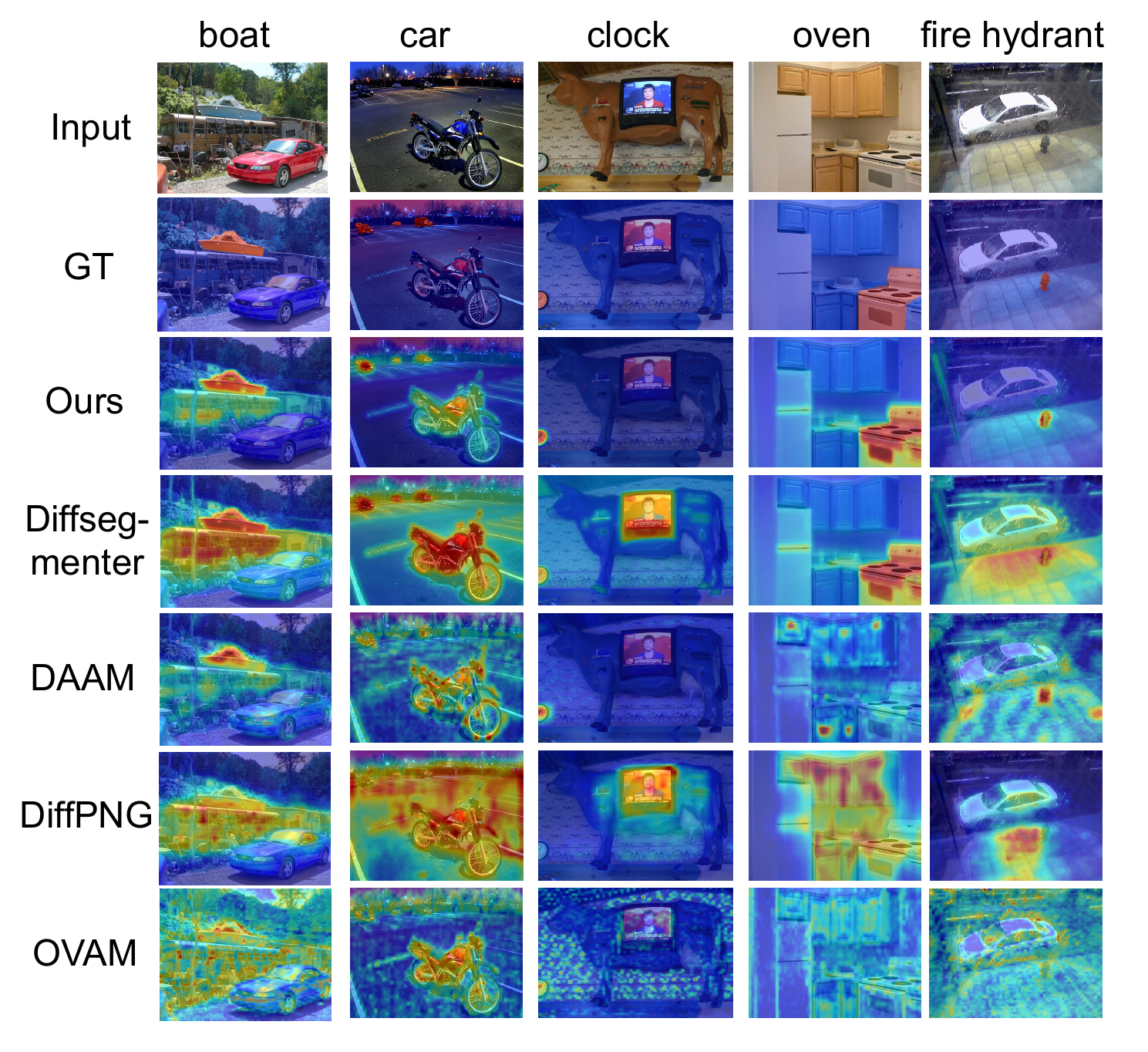}
    \caption{\textbf{Heatmaps generated by our proposed ELBO-T2IAlign and SOTA diffusion-based methods.}}
    \label{figure:method_heatmap}
\end{figure*}

\begin{figure*}[t]
    \centering
    \includegraphics[width=\linewidth]{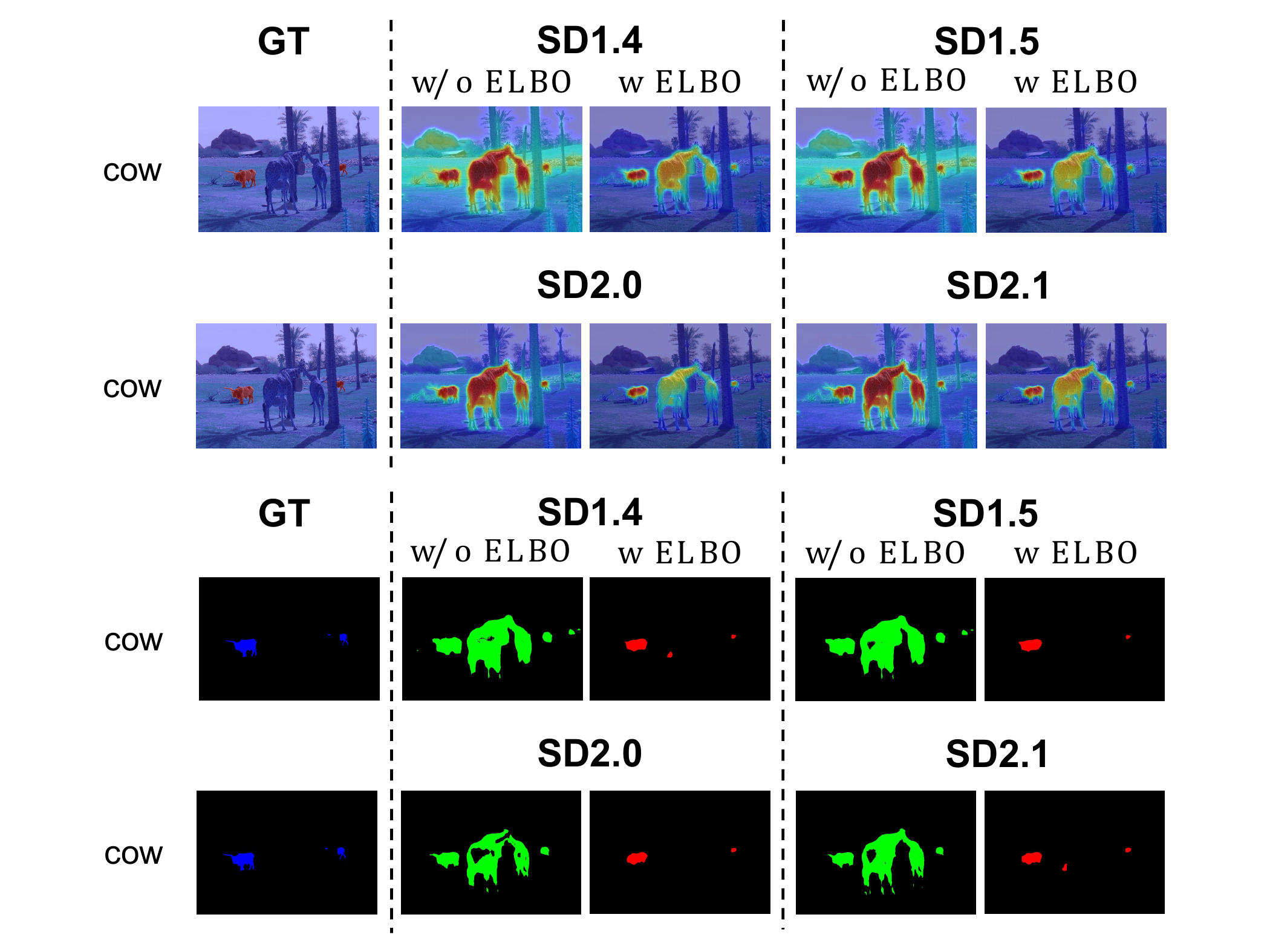}
    \caption{\textbf{Heatmaps and masks generated by different models with baseline and our ELBO-T2IAlign}}
    \label{figure:model_heatmap_mask}
\end{figure*}

In this section, we visualize more results of different methods and the performance of different versions of diffusion models under different $\gamma$ to demonstrate the effectiveness and generality of our approach. \figurename~\ref{figure:method_heatmap} presents a comparison of heatmaps generated by our method with those produced by other diffusion-based methods. \figurename~\ref{figure:model_heatmap_mask} illustrates the heatmaps and their corresponding masks generated by different versions of diffusion models with the baseline and our ELBO-T2IAlign.
These results further verify the effectiveness of the proposed method for pixel-text alignment.

\bibliographystyle{IEEEtran}
\bibliography{IEEEabrv,main}
\clearpage

